\pdfoutput=1

\documentclass[11pt]{article}

\usepackage[preprint]{acl}

\usepackage{times}
\usepackage{latexsym}
\usepackage{amsmath}
\usepackage{longtable}
\usepackage{colortbl}
\usepackage{amssymb}
\usepackage[T1]{fontenc}

\usepackage[utf8]{inputenc}

\usepackage{microtype}

\usepackage{inconsolata}
\usepackage{tcolorbox}
\usepackage{mdframed}

\usepackage{graphicx}
\usepackage{amssymb}
\usepackage{booktabs}
\usepackage{tabularx}
\usepackage{hyperref}
\usepackage{multirow}
\usepackage{subcaption}
\usepackage{cuted}
\usepackage{xtab}
\definecolor{lightgray}{RGB}{239,239,239}
\definecolor{mediumgray}{RGB}{204,204,204}
\definecolor{darkred}{HTML}{7e0f12}
\definecolor{darkgreen}{rgb}{0.0, 0.5, 0.0}
\definecolor{purple}{HTML}{7262ac}
\definecolor{softpink}{HTML}{FFD4DB}
\definecolor{softseafoam}{HTML}{ABD4D1}

\title{\raisebox{-0.09cm}{\includegraphics[width=0.55cm]{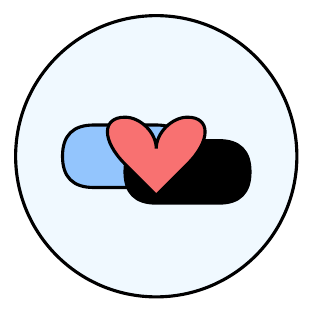}} What is Stigma Attributed to? A Theory-Grounded, Expert-Annotated Interview Corpus for Demystifying Mental-Health Stigma}

\author{
  Han Meng\textsuperscript{$\clubsuit$}, Yancan Chen\textsuperscript{$\clubsuit$}, Yunan Li\textsuperscript{$\clubsuit$}, Yitian Yang\textsuperscript{$\clubsuit$},\\
  \textbf{Jungup Lee\textsuperscript{$\diamondsuit$}, Renwen Zhang\textsuperscript{$\heartsuit$}, Yi-Chieh Lee\textsuperscript{$\clubsuit$}}\thanks{Corresponding author}
  \\
  \textsuperscript{$\clubsuit$}Department of Computer Science, National University of Singapore,\\
  \textsuperscript{$\diamondsuit$}Department of Social Work, National University of Singapore,\\
  \textsuperscript{$\heartsuit$}Department of Communications and New Media, National University of Singapore
  \\
  \texttt{\{han.meng, yancan, liyunan, yang.yitian\}@u.nus.edu},\\
  \texttt{\{swklj, r.zhang, yclee\}@nus.edu.sg}
}

\begin{document}
\maketitle
\begin{abstract}
\textcolor{red}{\textit{Warning: This paper contains content that may be offensive or disturbing, but this is unavoidable due to the nature of the work.}}

Mental-health stigma remains a pervasive social problem that hampers treatment-seeking and recovery. 
Existing resources for training neural models to finely classify such stigma are limited, relying primarily on social-media or synthetic data without theoretical underpinnings. 
To remedy this gap, we present an expert-annotated, theory-informed corpus of human-chatbot interviews, comprising 4,141 snippets from 684 participants with documented socio-cultural backgrounds. 
Our experiments benchmark state-of-the-art neural models and empirically unpack the challenges of stigma detection. 
This dataset can facilitate research on computationally detecting, neutralizing, and counteracting mental-health stigma. 
Our corpus is openly available at \url{https://github.com/HanMeng2004/Mental-Health-Stigma-Interview-Corpus}.
\end{abstract}

\section{Introduction}

\begin{figure}[t]
    \centering
    \includegraphics[width=0.985\linewidth]{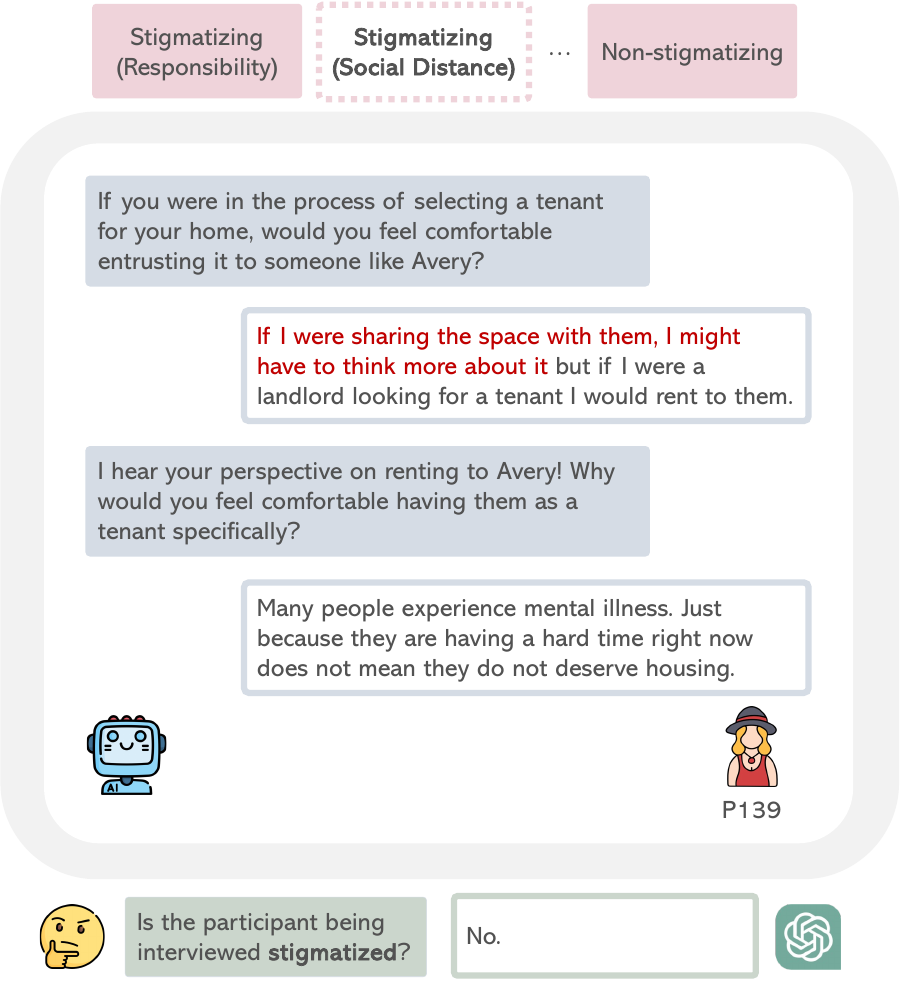}
    \caption{A sample interview snippet from our corpus where LLMs (here GPT-4o) fail to identify the underlying stigma, with the participant's response categorized as \textit{Stigmatizing (Social Distance)}. The stigmatizing text is highlighted in \textcolor{darkred}{\textbf{red}}.}
    \label{fig:overview}
\end{figure}

Mental illnesses profoundly shape the lives of hundreds of millions globally \cite{depression_prevalence_kessler_1994}. 
Yet those experiencing mental-health challenges often face a devastating double burden -- not just their symptoms, but also pervasive \textit{stigma} that leads to social isolation, workplace discrimination, and delayed treatment-seeking \cite{depression_stigma_peluso_2009}.
\textbf{Mental-health stigma}, as originally defined by Goffman, includes regarding mental illness as divergent from what society considers correct and normal, and mentally ill individuals as devalued and tainted \cite{stigma_spoiled_identity_goffman_1964}.
Social scientists have traditionally often qualitatively collected and coded texts from interviews that provide insider views to understand mental-health stigma, yet the considerable time, manual effort, and expertise required \cite{human_coding_bias_leeson_2019, coding_manual_saldana_2016} limit their breadth, fertility, and scalability.

Hence, the pressing need for a keystone dataset to train neural models capable of \textit{automatically} unraveling and disentangling stigma \cite{stigma_detect_giorgi_2024, llm_psycho_demszky_2023} has grown increasingly apparent, especially driven by the proliferation of digital mental-health discourse \cite{mental_health_discourse_garg_2022} and therapeutic conversational agents \cite{therapist_mishra_2023}.
However, unlike the growing body of research on general hate speech and offensive language targeting certain races and genders \cite{hate_speech_dataset_vidgen_2021, hate_speech_dataset_elSherief_2021, hate_speech_dataset_hartvigsen_2022, hate_speech_hamalainen_2021} (Table \ref{tab:dataset_comparison}), publicly available corpus specifically designed to understand mental-health stigma \cite{mental_health_stigma_corpus_mitkov_2023} remain scarce in the NLP community. 
More worryingly, these limited resources suffer from two critical limitations that presumably leave even state-of-the-art large language models (LLMs) not well-positioned to detect stigma effectively (Figure \ref{fig:overview}).

\begin{table*}[htbp]
  \centering
  \small
  \renewcommand{\arraystretch}{1.3}
  \begin{tabular}{l@{\hspace{5pt}}l@{\hspace{6pt}}r@{\hspace{6pt}}c@{\hspace{6pt}}c@{\hspace{6pt}}c@{\hspace{6pt}}c@{\hspace{6pt}}c}
    \toprule
    \textbf{Language Resource} & \textbf{Source} & \textbf{Size} & \textbf{Annotation} & \textbf{Research Scope} & \textbf{Public} & \textbf{Theory-} & \textbf{Socio-} \\[-2.8pt]
    & & & \textbf{Scheme} & & & \textbf{Grounded} & \textbf{cultural} \\
    \midrule
    \citet{hate_speech_dataset_elSherief_2021}
          & Twitter & 22,584 & Multi-Label & Hate Speech & \href{https://github.com/GT-SALT/implicit-hate}{\checkmark} & \checkmark & \\
    \citet{hate_speech_dataset_buyukdemirci_2024}
          & Twitter & 1,530 & Multi-Label & Hate Speech & \href{https://github.com/metunlp/JL-Hate}{\checkmark} &  & \\
    \citet{hate_speech_dataset_vidgen_2021} 
          & RoBERTa & 41,255 & Multi-Label & Hate Speech & \checkmark & & \\
    \citet{hate_speech_dataset_hartvigsen_2022}
          & GPT-3 & 274,186 & Binary & Hate Speech & \href{https://github.com/microsoft/ToxiGen}{\checkmark} & & \\
    \citet{offensive_dataset_baheti_2021} 
          & Reddit & 2,000 & Binary & Offensive Language & \href{https://github.com/abaheti95/ToxiChat}{\checkmark} & & \\
    \citet{stereotype_dataset_cignarella_2024} 
          & Facebook & 2,888 & Binary & Social Stereotypes & \href{https://github.com/bvidgen/Dynamically-Generated-Hate-Speech-Dataset}{\checkmark} & & \\   
    \citet{microaggression_dataset_breitfeller_2019}
          & Tumblr & 2,934 & Multi-Label & Microaggression & \href{https://bit.ly/2lVv3BG}{\checkmark} & \checkmark & \\
 \citet{stigma_detect_straton_2020}
          & Facebook & 2,761 & Multi-Label & Vaccine Stigma & & \checkmark & \\
    \citet{stigma_detect_roesler_2024}
    & Reddit & 2,214 & Multi-Label & Substance Use Stigma\footnotemark & & \checkmark & \\
    \citet{mental_health_stigma_corpus_mitkov_2023}
          & ChatGPT & 9,700 & Multi-Label & Mental-Health Stigma & \href{https://github.com/masonchoey/from-stigma-to-support}{\checkmark} & & \\
    \textsc{MHStigmaInterview} (ours) 
          & Interview & 4,141 & Multi-Label & Mental-Health Stigma & \checkmark & \checkmark & \checkmark \\
    \bottomrule
  \end{tabular}
  \caption{Summary and comparison of our corpus with existing datasets on problematic language and mental-health stigma. \textit{Annotation Scheme} indicates whether a dataset uses binary or multi-class labeling. \textit{Theory-Grounded} shows if the annotations draw from sociological and/or psychological theories. \textit{Socio-cultural} indicates whether a dataset records the socio-cultural background of its data contributors/sources. Ours is the first interview-based corpus for mental-health stigma.}
  \label{tab:dataset_comparison}
\end{table*}

First, current corpora are almost \textbf{exclusively sourced from social-media platforms} \cite{detect_method_jilka_2022, stigma_detect_straton_2020}, where discourse tends to be polarized and inflammatory \cite{polarize_social_media_moriceau_2022, social_media_decontext_boyd_2012}, or from \textbf{synthetic} data \cite{mental_health_stigma_corpus_mitkov_2023}.
Interview data analyzed by social scientists, on the other hand, are renowned for interpersonal, conversational context and rich human narratives, their capacity to elicit self-disclosure and deep reflection \cite{conversation_importance_jenlink_2005, self_disclosure_ho_2018}, as well as being rife with psychological constructs, perceptions, and valuable information about how social problems are negotiated, unfolded, and perpetuated \cite{interview_stigma_measure_liggins_2005, interview_stigma_measure_lyons_1995}.
Such a mismatch between computational and social-science data sources puts neural classifiers at risk of over-fitting to certain lexical and linguistic markers, such as swear words and Internet buzzwords \cite{mental_health_social_media_harrigian_2020}.

\footnotetext{Substance use disorder is classified as a mental illness in DSM-5 \cite{dsm5_apa_2013}. We follow \citet{attribution_model_corrigan_2003} in using the term \textit{mental-health stigma}.}

Another potential drawback -- particularly critical given that socio-cultural factors are deeply associated and intertwined with mental-health stigma \cite{culture_difference_krendl_2020} -- is that, unlike interview participants whose demographic backgrounds can be systematically documented, the texts from social media and synthetic corpora largely come with little to no information about who created them and the social and cultural contexts of those creators \cite{social_media_deidentify_ruths_2014}.

Second, integration with \textbf{psychological and sociological theories} of mental-health stigma in existing datasets ranges from simplistic \cite{stigma_detect_roesler_2024} to nonexistent.
Such social models and conceptual frameworks are essential for an empirical, fine-grained understanding of stigmatizing language \cite{theory_nlp_bonikowski_2022, theory_nlp_hovy_2021, social_factor_language_blodgett_2020}. 
Their absence has led to computational methods that are largely confined to binary classification \cite{detect_method_jilka_2022, detect_method_lee_2022} and are unlikely to capture the full spectrum of psycholinguistic nuances of stigmatization -- from the casual appropriation of diagnostic terms \cite{appropriation_moriceau_2022} to euphemisms \cite{euphemistic_magu_2018} and microaggressions \cite{microaggression_dataset_breitfeller_2019}.

Motivated by these two critical gaps and informed by prior work on chatbots' capacity to conduct interviews \cite{chatbot_aq27_practice_lee_2023, disclosure_lee_2022}, we introduce a new corpus of interviews collected by chatbots and annotated using a protocol guided by the \textit{attribution model} \cite{attribution_model_corrigan_2003} -- a widely-adopted theoretical framework for deconstructing mental-health stigma.
This rich corpus adds value to the NLP community in three ways: 1) to the best of our knowledge, it represents the first large-scale, open-source mental-health stigma interview dataset, comprising 4,141 transcript snippets from 684 human participants; 2) it captures real-world interactions with clear socio-cultural representation and rich contextual information, thereby contributing fresh linguistic resources for understanding stigmatizing language; and 3) it features expert-led, theoretically-grounded labels that help equip neural models with the ability to identify fine-grained drivers of stigmatization.
Finally, we fine-tune classifiers like RoBERTa \cite{roberta_liu_2019} and prompt LLMs such as Llama \cite{llama3_dubey_2024} and GPT-4o for stigma detection, providing both useful benchmarks and empirical insights to sustain future research.

\section{Background and Related Work}

Problematic language and social bias have long been central concerns in NLP research \cite{hate_speech_detect_schmidt_2017, hate_speech_rottger_2021, hate_speech_dataset_lee_2024}.
From offensive and abusive language laced with profanity \cite{hate_speech_sanguinetti_2018} to toxic and derogatory hate speech that disparages people based on their protected characteristics and social identities \cite{hate_speech_vargas_2022}, to outright hostile rhetoric and trolling \cite{troll_lee_2022} that fuels online harassment \cite{harassment_liu_2019}, to seemingly mild but harmful comments that breed disrespect \cite{hate_speech_lu_2023}, researchers have examined how to classify and mitigate it \cite{hate_speech_korre_2024}.\footnote{See Appendix \ref{app:term} for a discussion of the definitions of these harmful language categories.} 
Several influential benchmarks and language resources, summarized in Table \ref{tab:dataset_comparison}, tremendously help detect and identify these harmful languages: for example, \citet{hate_speech_dataset_elSherief_2021} tackles white grievance, incitement to violence, expressions of perceived inferiority, and so on; \citet{hate_speech_dataset_buyukdemirci_2024} annotates hate-speech targets and signals at the token level; \citet{hate_speech_dataset_vidgen_2021} examines dehumanizing language, threats, and displays of animosity; and \citet{microaggression_dataset_breitfeller_2019} explores microaggressions against certain genders, religions, and ages in everyday discourse.

Yet, we consider that these carefully curated and compiled resources are not sufficient to automatically decipher mental-health stigma. 
One obstacle is that these corpora not only lack coverage of people with mental illness as a target population \cite{hate_target_yu_2024}, but more fundamentally, mental-health stigma is uniquely operationalized through cognitive judgments of personal flaws that drive negative \textit{emotions} and \textit{behavioral} intentions \cite{attribution_theory_corrigan_2000} -- requiring the capture of both linguistic markers and underlying attributional chains \cite{attribution_model_corrigan_2003}.
Given this complexity, it is concerning that the sole open-source corpus in this space \cite{mental_health_stigma_corpus_mitkov_2023}, shown in Table \ref{tab:dataset_comparison}, presumably suffers from homogeneous \textbf{psycholinguistic patterns} \cite{decontextualize_patton_2020, social_media_decontext_boyd_2012}, with limited awareness of socio-cultural factors \cite{culture_difference_krendl_2020, social_media_deidentify_ruths_2014} and potentially \textbf{questionable theoretical grounding} \cite{theory_nlp_hovy_2021}.

To address this resource gap, and to further answer the recent calls for NLP scholars to invest in keystone datasets \cite{llm_psycho_demszky_2023}, integrate psychological theories \cite{theory_nlp_bonikowski_2022, theory_nlp_boyd_2021}, better understand human behavior through language \cite{theory_mihalcea_2024}, and develop socially aware technologies that carefully consider how social factors shape language use \cite{theory_yang_2024}, we provided a public, large-scale collection of interview snippets with clear demographic documentation, where we also break down theoretical constructs into specific, measurable categories, namely labels.

\section{Data}

\begin{table*}[tbp]
\small
\renewcommand{\arraystretch}{1.3}
\begin{tabular}{p{0.45\textwidth}p{0.32\textwidth}p{0.1\textwidth}}
\toprule
\multicolumn{1}{c}{\textbf{Interview Question in \textsc{MHStigmaInterview}}} & \multicolumn{1}{c}{\textbf{AQ-27 Survey Item \cite{aq27_corrigan_2012}}} & \multicolumn{1}{c}{\textbf{Attribution Type}}\\
\midrule
Do you believe Avery's current situation is primarily \textbf{a result of their actions}? Please share your thoughts. & I would think that it was Harry's \textbf{own fault} that he is in the present condition. & Responsibility \\
If you were in the process of selecting a \textbf{tenant} for your home, would you feel comfortable entrusting it to someone like Avery? & If I were a \textbf{landlord}, I probably would rent an apartment to Harry. & Social Distance \\
Let's imagine you were one of Avery's friends at the party or other social occasions. Would you feel \textbf{anger} toward them due to their actions, for example, losing their temper and yelling at someone? & How \textbf{angry} would you feel at Harry? & Anger \\
I wonder if you were one of Avery's colleagues or friends, then would you be inclined to \textbf{help} them with tasks, such as work projects? & How likely is it that you would \textbf{help} Harry? & Helping \\
I'd like to know if you would feel \textbf{concern} and \textbf{sympathy} at Avery's mental state? & How much \textbf{sympathy} and \textbf{concern} would you feel for Harry? & Pity \\
If you are Avery's neighbor, then would you consider it better to \textbf{separate} them from the neighborhood and have them undergo \textbf{hospitalization} for treatment? & How much do you think an \textbf{asylum}, where Harry can be \textbf{kept away from} his neighbors, is the best place for him? & Coercive Segregation \\
If you were traveling or engaging in activities with Avery, would you have any feelings of \textbf{frightened} or being \textbf{threatened} regarding the potential implications of your interaction? & I would feel \textbf{frightened} and \textbf{threatened} by Harry. & Fear \\
\bottomrule
\end{tabular}
\caption{Interview questions and their corresponding theoretical underpinnings.}
\label{tab:question}
\end{table*}

\subsection{Theoretical Framework}
\label{sec:theory}

\paragraph{\textcolor{darkred}{\textit{Attribution Theory.}}}
Attribution theory \cite{attribution_theory_kelley_1967} is a core psychological concept that examines how individuals interpret and assign causes to events and behaviors, particularly whether they attribute outcomes to "internal" (dispositional) or "external" (situational) factors. 
Our corpus curation is informed by \citet{attribution_model_corrigan_2003}'s \textit{attribution model}, an established attribution theory that deconstructs public mental-health stigma in terms of \textit{cognitive judgments} (i.e., stereotypes), \textit{emotional responses} (i.e., prejudice), and \textit{behavioral reactions} (i.e., discrimination). 
Specifically, the model posits that people's perceptions of \textit{responsibility} for mental illness lead to emotions such as \textit{anger}, \textit{fear}, and/or \textit{pity}, which in turn drive discriminatory behaviors like \textit{avoidance}, \textit{coercive segregation}, and/or withholding \textit{help}. 

\paragraph{\textcolor{darkred}{\textit{Operationalization.}}}
Under the guidance of the mental-health specialist (a co-author), we operationalized the theoretical constructs by adapting the Attribution Questionnaire-27 (AQ-27) \cite{aq27_corrigan_2012}, a standardized survey developed from the attribution model that decomposes mental-health stigma into nine measurable items -- \textit{blame} (i.e., responsibility), \textit{fear}, \textit{pity}, \textit{anger}, \textit{helping}, \textit{avoidance} (i.e., social distance), \textit{coercion}, \textit{segregation}, and \textit{dangerousness}. 
Notably, following \citet{chatbot_aq27_practice_lee_2023} and \citet{aq_27_practise_meng_2024}, we combined the fear-dangerousness and coercion-segregation pairs to reduce repetitiveness and create a more concise interview, yielding seven key attributions that served as the basis for both our \textbf{interview protocol} (Table \ref{tab:question}) and \textbf{annotation scheme}. 
The definitions of these attributions appear in Appendix \ref{app:attributions}.

\subsection{Data Collection}

\subsubsection{Chatbot-based Interview Design}

\paragraph{\textcolor{darkred}{\textit{Interview Flow.}}}

We designed and programmed a chatbot via UChat\footnote{\url{https://uchat.au/}} to conduct 20-minute dyadic interactions with participants, with our interview structure informed by prior work \cite{vignette_alem_1999, aq_27_practise_meng_2024, chatbot_aq27_practice_lee_2023}. 

First, the chatbot initiated a \textbf{rapport-building small-talk session} \cite{small_talk_bockmore_1999}, discussing topics such as participants' favorite movies and activities. 
Then, it presented a \textbf{vignette} \cite{vig_lerner_2010} about "\textit{Avery}," a fictional character of unspecified demographics who suffers from a major depressive disorder \cite{dsm5_apa_2013} affecting their academic performance, work productivity, and social relationships. 
And finally, the chatbot engaged in the \textbf{question-answer} conversation -- each interview question (Table \ref{tab:question}) was embedded within a vivid, relatable scenario based on Avery's vignette to mitigate social-desirability bias \cite{sd_indirect_q_fisher_1993, sd_scale_inaccurate_van_2008}, and the questions were presented in randomized order to avoid priming effects \cite{priming_effect_molden_2014}.

All materials emanating from the chatbot underwent expert review by a consulting psychiatrist and the mental-health specialist. 
It should be noted that our released corpus excludes the initial small-talk and vignette delivery, as these sections primarily serve as conversation warm-ups, fall outside our annotation scope, and/or may contain sensitive personal information.
Further details about the interview can be found in Appendix \ref{app:collection}.

\paragraph{\textcolor{darkred}{\textit{Vignette.}}}

\textit{Vignettes} serve as powerful research instruments for examining attitudes about mental health through brief fictional stories \cite{vignette_alem_1999}. 
Based on research findings and lived experiences \cite{chatbot_aq27_practice_lee_2023, vignette_design_griffiths_2006}, these narratives enable participants to react to specific situations, giving researchers deeper insights into their views. 
Specifically, Avery's symptoms were set forth in the DSM-5 \cite{dsm5_apa_2013}, though we excluded more severe manifestations, such as self-harm and suicidal behavior, and we avoided medical and/or technical jargon.
Two versions were developed: a \textbf{clinical version} that systematically describes Avery's symptoms and a \textbf{narrative version} that brings the story to life with dialogue, actions, and vivid details. 
Participants only saw the narrative version as the chatbot delivered it sentence by sentence, whereas researchers, including annotators, had access to \textbf{both} versions.
Both vignette versions appear in Appendix \ref{app:collection}.

\paragraph{\textcolor{darkred}{\textit{Interview Questions and their Follow-up Question(s).}}}

Our chatbot administered seven core \textbf{interview questions} ($\mathcal{IQ}s$) aligned with the attributions in Table \ref{tab:question}, supplemented by \textbf{follow-up questions} ($\mathcal{FQ}s$) \cite{follow_up_q_han_2021} to foster self-disclosure.
All $\mathcal{FQ}s$ (if needed) are specifically designed for certain $\mathcal{IQ}s$ and are completed before moving on to the next $\mathcal{IQ}$.

Specifically, we designed our questioning protocol $q(r)$ based on the length of participants' responses $r$:

\begin{align}
\small
q(r) = \begin{cases} 
\mathcal{IQ} + \mathcal{FQ}_{\text{1}} + \mathcal{FQ}_{\text{2}}, & \begin{array}{l} 
    \text{if } |r| < 25 \text{ and} \\
    |r + \mathcal{FQ}_{\text{1}}| < 150
\end{array} \\[3ex]
\mathcal{IQ} + \mathcal{FQ}_{\text{1}}, & \begin{array}{l}
    \text{if } |r| < 25 \text{ and} \\
    |r + \mathcal{FQ}_{\text{1}}| \geq 150
\end{array} \\[3ex]
\mathcal{IQ} + \mathcal{FQ}_{\text{2}}, & \text{if } 25 \leq |r| \leq 150 \\[2ex]
\mathcal{IQ}, & \text{if } |r| > 150
\end{cases}, 
\end{align}

where $|r|$ denotes the length of the response to $\mathcal{IQ}s$ in characters, and $\mathcal{FQ}s$ are specific to each attribution -- for \textit{responsibility} attribution, $\mathcal{FQ}1$ explores underlying \textbf{reasons} and $\mathcal{FQ}2$ examines if participants view mental illness as \textbf{personal weakness}.\footnote{This $\mathcal{FQ}$ was derived from \citet{stigma_reduce_cui_2024} and reviewed and validated by the mental-health specialist.} 
For \textit{emotional responses}, $\mathcal{FQ}1$ similarly probes \textbf{reasons} while $\mathcal{FQ}2$ asks participants to identify specific aspects of Avery's story that evoked their emotions. 
For \textit{behavioral responses}, $\mathcal{FQ}1$ takes two forms: asking about \textbf{potential outcomes} for stigmatizing responses ($\mathcal{FQ}1_a$) or exploring \textbf{reasons} for non-stigmatizing ones ($\mathcal{FQ}1_b$), followed by $\mathcal{FQ}2$ about triggering vignette plots.\footnote{The two thresholds were determined through an 8-participant pilot study and in consultation with the mental-health specialist.}

\subsubsection{Participant Recruitment}

We amassed participants through the research platforms Prolific\footnote{\url{https://www.prolific.com/}} and Qualtrics.\footnote{\url{https://www.qualtrics.com/}} 
Specifically, participants were required to meet the following inclusion criteria: 1) be at least 21 years of age, 2) have English as their first language, 3) be willing to engage with content related to mental illness, and 4) have no immediate or pressing mental-health concerns, as assessed using the Kessler Screening Scale for Psychological Distress (K6) \cite{k6_kessler_2003}.
We included this last criterion to safeguard vulnerable individuals from potential distress and trauma when being exposed to depression-related vignettes \cite{ethic_mental_illness_roberts_2002}. 

Our recruitment materials clearly outlined the duration and scope of the study and participants' right to withdraw.
We collected responses from 684 participants, using IP verification to prevent duplicates, with demographics detailed in Appendix \ref{app:demographic}.

\subsubsection{Consent and Procedure}

This 30-minute, single-session study compensated participants US\$6.30, following both platform-standard rates\footnote{\url{https://researcher-help.prolific.com/en/article/9cd998}} and the Department Ethics Review Committee (DERC) guidelines at the main researcher's institution. 

Participants began by receiving a warning about the interview's mental health focus. 
They then reviewed and completed the Participant Information Statement and Consent Form (PISCF) \cite{consent_nijhawan_2013}, which detailed privacy protocols and how their data would be collected, stored, and used in our research.
Upon obtaining their consent, we reminded participants of their right to withdraw and invited them to voluntarily share demographic information. 
The core data-collection session consisted of a 20-minute interaction with the chatbot, which concluded with a \textit{debriefing} \cite{debriefing_fanning_2007} that covered our research objectives and common misconceptions about mental health.

\subsection{Data Annotation and Filtering}

\paragraph{\textcolor{darkred}{\textit{Annotation Setup.}}}

We opted for \textbf{expert-guided annotation} \cite{expert_annotate_snow_2008} over crowdsourcing -- given 1) the need for a deep understanding of social-psychological theories \cite{attribution_model_corrigan_2003} (akin to qualitative coding \cite{coding_manual_saldana_2016}) and domain knowledge, 2) the potential negative impact of reviewing stigmatized texts on annotators' mental well-being, and 3) the importance of close monitoring and sustained discussion.
Specifically, with positionality and triangulation strategies \cite{hci_qualitative_coding_method_lazar_2017} in mind, we hired two full-time research assistants (RAs) for a three-month period: a computer science-trained Asian male and a social science-educated Asian female, both in their twenties.
The RAs annotated under the joint guidance of the mental-health specialist and the main researcher, with compensation following institutional regulations.

Each data unit consisted of an interview snippet centered on one attribution. 
Two annotators independently reviewed each snippet and the accompanying annotation instructions on the \textsc{Potato} platform \cite{potato_pei_2022}, answered a \textbf{multiple-choice question} (i.e., "\textit{Does this interview snippet contain stigma, and if identified, which specific attribution type is present?}"), and selected one of the seven attributions described in Section \ref{sec:theory} or a "\textit{Non-stigmatizing}" option. 
They could also mark snippets as "\textit{N/A}" when participant responses 1) provided minimal information (e.g., only "\textit{yes}," "\textit{no}," or "\textit{not sure}"), 2) contained only irrelevant content, 3) exhibited evident AI-generated characteristics, and/or 4) were incomplete or indecipherable.
The annotation platform interface, examples, and instructions are provided in Appendix \ref{app:annotation}.

\paragraph{\textcolor{darkred}{\textit{Annotation Process.}}}

The team first developed an initial codebook (i.e., annotation instructions) based on the attribution model \cite{attribution_model_corrigan_2003}, which included definitions adapted from \citet{aq27_corrigan_2012}, keywords, examples, and rules/specifications for each label.
Following the mental-health specialist's guidance, we refined the codebook through iterative revisions and practice rounds until the agreement, as measured by \textbf{Cohen's $\kappa$} \cite{cohens_kappa_mchugh_2012}, reached 0.6 \cite{coding_manual_saldana_2016}, at which point we finalized the codebook.
We established checkpoints to assess inter-rater agreement: starting with two batches of 10 participants ($\kappa=0.55$, $0.53$), moving to four sets of 20 participants ($\kappa=0.66$, $0.79$, $0.76$, and $0.72$). 
As consistency improved, we progressed to larger samples with two sets of 50 participants ($\kappa=0.74$, $0.66$), one set of 100 participants ($\kappa=0.69$), one set of 150 participants ($\kappa=0.69$), and the remaining participants ($\kappa=0.66$). 

At each checkpoint, the team held regular meetings to openly discuss and resolve any disagreements.
We excluded interview snippets that 1) were consistently labeled "\textit{N/A}" by both annotators, and 2) were labeled differently by the two annotators and later unanimously agreed to be "\textit{N/A}" through open discussion. 
The final corpus comprised 4,141 interview snippets (86.49\% of the original data), with annotators achieving both a satisfactory $\kappa$ of 0.71 \cite{cohens_kappa_mchugh_2012} and perfect agreement on 87.42\% of all snippets.
Appendix \ref{app:matrix} presents the agreement matrix between two annotators.

\subsection{Corpus Statistics}

In this section, we analyze key statistical features of our expert-annotated corpus. 
The collected interview snippets average 2.11 turns ($SD = 0.63$) between participants and the chatbot, with Table \ref{tab:label-distribution} depicting that responses indicating personal \textbf{responsibility} and social \textbf{avoidance} emerge as the most common stigmatization.

Specifically, an interview snippet annotated as \textit{Stigmatizing (responsibility)} is shown below:

\begin{quote}
\textbf{Chatbot}: \textit{Do you believe Avery's current situation is primarily a result of their actions? Please share your thoughts.}\\
\textbf{Participant}: \textit{I do think its partially}\\
\textbf{Chatbot}: \textit{That's a really fair perspective to have! What's your reason behind it?}\\
\textbf{Participant}: \textit{We don't know his backstory, how his family life is, anything that's happened is probably not so good.}\\
\textbf{Chatbot}: \textit{That makes a lot of sense; understanding someone's background is crucial. Do you think Avery's mental condition reflects personal weakness, or is it more a response to challenging life circumstances?}\\
\textbf{Participant}: \textit{I do not, mental issues are never a personal weakness, you can prove you're strong by taking control of it.}
\end{quote}

In contrast, the following is a stigma-free interview snippet annotated as \textit{Non-stigmatizing}:

\begin{quote}
\textbf{Chatbot}: \textit{If you are Avery's neighbor, would you consider it better to separate them from the neighborhood and have them undergo hospitalization for treatment?}\\
\textbf{Participant}: \textit{No, I don't think that's the way to do it. She could seek help from a medical professional, but hospitalization seems extreme.}\\
\textbf{Chatbot}: \textit{Got it, you feel hospitalization might be too extreme. What specific plot points in Avery's story made you think seeking help, but not hospitalization, would be a better route?}\\
\textbf{Participant}: \textit{She is suffering from depression and outbursts. I don't think that qualifies to be hospitalized.}
\end{quote}

\begin{figure*}[t]
    \centering
    \includegraphics[width=\textwidth]{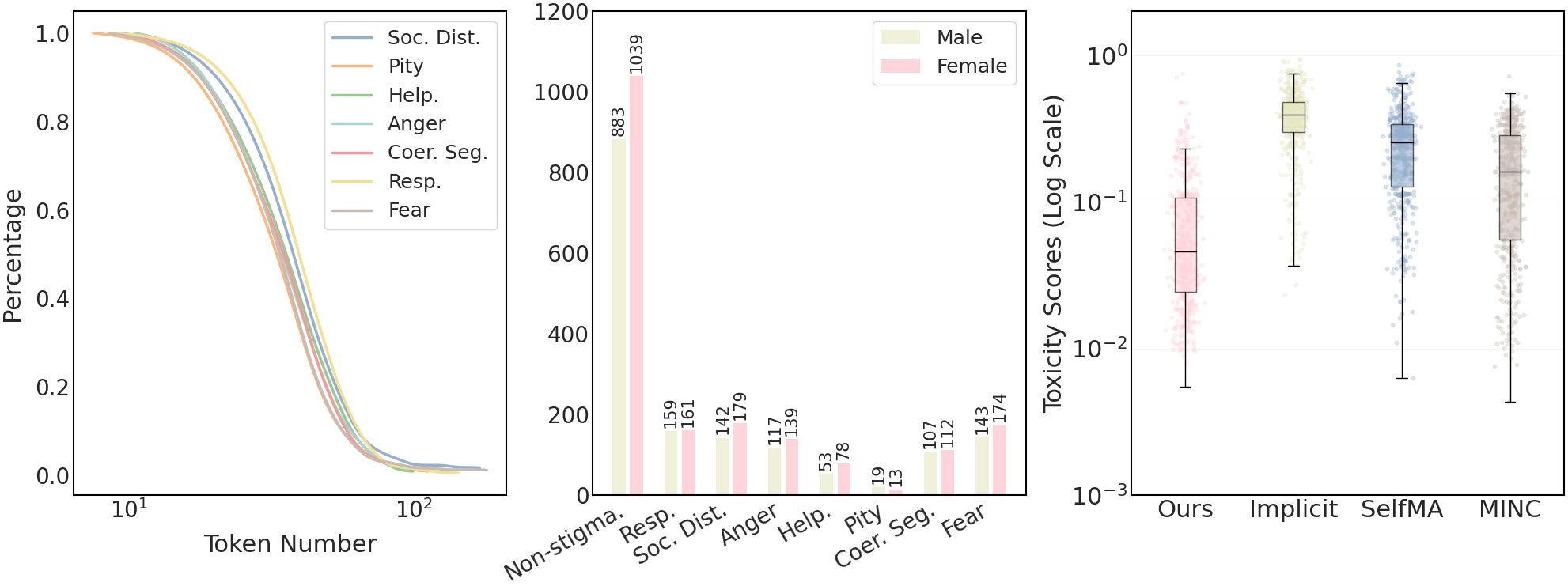}
\caption{Token-count distribution across attributions (left), the association between mental-health stigma and participants' socio-demographic background (e.g., gender) (center), and toxicity-score comparison between our corpus and three benchmark datasets (right).}
    \label{fig:stats}
\end{figure*}

The token-count analysis (Figure \ref{fig:stats}) shows longer, more elaborate responses to personal blame ($M = 41.97$, $SD = 17.50$), presumably reflecting participants' internal conflicts about individual agency versus systemic factors in making moral judgments about mental illness. 
In addition, our preliminary analyses of socio-cultural factors \cite{culture_difference_krendl_2020} suggest that stigmatizing attribution is partly intertwined with interlocutors' demographic backgrounds and geographic locations (Figure \ref{fig:stats} and \ref{fig:socio}).
Additional analyses of correlations between different stigma attributions and emotions, along with response-quality patterns, are detailed in Appendix \ref{app:more_stat}.

\begin{table}[tbp]
\small
\centering
\begin{tabular}{lrr}
\toprule
Label & \# Snippet &  \% \\
\midrule
Non-stigmatizing & 2,232 & 53.90 \\
Stigmatizing & - & - \\ 
\hspace{0.5cm}Responsibility & 394 & 9.51 \\
\hspace{0.5cm}Social Distance & 379 & 9.15 \\
\hspace{0.5cm}Anger & 298 & 7.20 \\
\hspace{0.5cm}Helping & 158 & 3.82 \\
\hspace{0.5cm}Pity & 42 & 1.01 \\
\hspace{0.5cm}Coercive Segregation & 271 & 6.54 \\
\hspace{0.5cm}Fear & 367 & 8.86 \\
\midrule
Total & 4,141 & 100 \\
\bottomrule
\end{tabular}
\caption{Label distribution in our corpus, including a non-stigmatizing category and seven stigma attributions operationalized from the attribution model.}
\label{tab:label-distribution}
\end{table}

To situate our corpus, we compared it with three datasets of problematic language listed in Table \ref{tab:dataset_comparison}.\footnote{We extracted and analyzed texts annotated as microaggressions in \citet{microaggression_dataset_breitfeller_2019} and those labeled as \textit{implicit} hate speech in \citet{hate_speech_dataset_elSherief_2021}.} 
As shown in Figure \ref{fig:more_stats}, the semantic distributions across these datasets diverge markedly, presumably due to our unique interview-driven discourse genre and our focal target of mentally ill individuals, an understudied minority, thus adding unique value to existing language resources in the NLP community.
Surprisingly, the stigmatizing texts in our corpus show even \textbf{lower toxicity scores}\footnote{Calculated by Perspective API (\url{https://perspectiveapi.com/}).} than content from those benchmark datasets (Figure \ref{fig:stats}) -- despite their reputation for capturing veiled, oblique, and understated forms of hate -- illuminating increasingly elusive, insidious instances of stigma embedded in richly contextualized interactions. 
For instance, responses labeled as \textit{Stigmatizing (Helping)} display toxicity scores indistinguishable from non-stigmatizing texts (Figure \ref{fig:more_stats}), exemplifying how stigmatization can be interwoven into neutral or even ostensibly benign discussions.

\section{Experiments: Mental-Health Stigma Detection}

\subsection{Experimental Setup}

To benchmark how well computational models can detect mental-health stigma, we set up an 8-way classification task, i.e., \textbf{stigma detection}, using our interview corpus. 
Specifically, we experimented with state-of-the-art LLMs, including GPT-4o (transformer-based, instruction-tuned) \cite{gpt4_achiam_2023}, Llama-3.1-8B (decoder-only transformer, instruction-tuned), Llama-3.3-70B (decoder-only transformer, instruction-tuned) \cite{llama3_dubey_2024}, Mistral Nemo (\texttt{Mistral-Nemo-Instruct-2407}, decoder-only transformer, instruction-tuned), and Mixtral 8×7B (\texttt{Mixtral-8x7B-Instruct-v0.1}, decoder-only mixture-of-experts transformer, instruction-tuned) \cite{mixtral_jiang_2024}, alongside a fine-tuned RoBERTa-base model (encoder-only transformer, base model) \cite{roberta_liu_2019}. 
We sampled 60\%, 20\%, and 20\% of instances from each category to create the train, validation, and test splits. 
The experiments ran on a cluster of four H100 GPUs for approximately 150 hours, with an additional US\$500 allocated for GPT-4o API calls.

For the hyperparameter search, we explored temperature values between \{0.0, 0.1, 0.2, 0.3, 0.4\} and selected 0.2 as our best-found setting for Llama-3.1-8B, Llama-3.3-70B, and Mixtral 8×7B; 0.3 was set for Mistral Nemo following its documentation recommendations, while GPT-4o was only tested at 0.2 due to computational budget constraints. 
Next, we evaluated epochs \{2, 3, 4\} and learning rates \{5e-5, 1e-5\} for fine-tuning RoBERTa, with 4 epochs and a learning rate of 5e-5 being the best-found configuration, along with the AdamW optimizer and a batch size of 12. 
We implemented \texttt{FlashAttention-2} \cite{flashattention_dao_2024} to improve efficiency. 
We averaged the results over three runs for all models except GPT-4o (which had a single run due to cost constraints) to account for randomness.

We experimented with three prompt settings: zero-shot, one-shot, and \textbf{full codebook} information matching the guidelines given to human annotators (full prompts available in Appendix \ref{app:prompt}).

\begin{table*}[ht]
\small
\centering

\renewcommand{\arraystretch}{1.15} 

\resizebox{\linewidth}{!}{%
\begin{tabular}{>{\centering\arraybackslash}llcccccccccc} 
    \toprule
    \multirow{2}{*}[-0.5ex]{\textbf{Model}} 
    & \multicolumn{5}{c}{\textbf{Zero-shot}} 
    & \multicolumn{5}{c}{\textbf{One-shot}} \\
    \cmidrule(lr){2-6} \cmidrule(lr){7-11}
    & \textit{P} & \textit{R} & \textit{F1} & \textit{Cohen's $\kappa$} & \textit{Acc} 
    & \textit{P} & \textit{R} & \textit{F1} & \textit{Cohen's $\kappa$} & \textit{Acc} \\
    \midrule
    GPT-4o        & .572 & .446 & \colorbox[rgb]{1.0, 0.875, 0.894}{.456} & \colorbox[rgb]{1.0, 0.875, 0.894}{.394} & .578  
                  & \colorbox[rgb]{1.0, 0.875, 0.894}{.590} & \colorbox[rgb]{1.0, 0.875, 0.894}{.737} & \colorbox[rgb]{1.0, 0.875, 0.894}{.576} & \colorbox[rgb]{1.0, 0.875, 0.894}{.473} & \colorbox[rgb]{1.0, 0.875, 0.894}{.561} \\
                  
    Llama-3.1-8B   & .479 & .445 & .307 & .206 & .267  
                  & .493 & .555 & .383 & .260 & .306 \\

    Llama-3.3-70B  & \colorbox[rgb]{1.0, 0.875, 0.894}{.610} & \colorbox[rgb]{1.0, 0.875, 0.894}{.447} & .449 & .311 & .425  
                  & .581 & .648 & .545 & .416 & .505 \\

    Mistral Nemo  & .356 & .266 & .238 & .278 & .350  
                  & .533 & .645 & .478 & .389 & .473 \\

    Mixtral 8×7B   & .439 & .304 & .318 & .379 & \colorbox[rgb]{1.0, 0.875, 0.894}{.608}  
                  & .471 & .503 & .429 & .335 & .504 \\

    RoBERTa       & —    & —    & —    & —    & —    
                  & —    & —    & —    & —    & — \\
    \bottomrule
\end{tabular}%
}

\vspace{1pt} %

\resizebox{\linewidth}{!}{%
\begin{tabular}{>{\centering\arraybackslash}llcccccccccc}
    \toprule
    \multirow{2}{*}[-0.5ex]{\textbf{Model}}
    & \multicolumn{5}{c}{\textbf{Full Codebook}} 
    & \multicolumn{5}{c}{\textbf{Fine-tune}} \\
    \cmidrule(lr){2-6} \cmidrule(lr){7-11}
    & \textit{P} & \textit{R} & \textit{F1} & \textit{Cohen's $\kappa$} & \textit{Acc} 
    & \textit{P} & \textit{R} & \textit{F1} & \textit{Cohen's $\kappa$} & \textit{Acc} \\
    \midrule
    GPT-4o        & .742 & \colorbox[rgb]{1.0, 0.875, 0.894}{.801} & \colorbox[rgb]{1.0, 0.875, 0.894}{.757} & .763 & .835  
                  & —    & —    & —    & —    & — \\

    Llama-3.1-8B   & .545 & .732 & .521 & .407 & .472  
                  & —    & —    & —    & —    & — \\

    Llama-3.3-70B  & \colorbox[rgb]{1.0, 0.875, 0.894}{.778} & .744 & .752 & \colorbox[rgb]{1.0, 0.875, 0.894}{.767} & \colorbox[rgb]{1.0, 0.875, 0.894}{.847} 
                  & —    & —    & —    & —    & — \\

    Mistral Nemo  & .643 & .779 & .662 & .620 & .708  
                  & —    & —    & —    & —    & — \\

    Mixtral 8×7B   & .660 & .602 & .584 & .552 & .705  
                  & —    & —    & —    & —    & — \\

    RoBERTa       & —    & —    & —    & —    & —    
                  & .747 & .766 & .885 & .755 & .832 \\
    \bottomrule
\end{tabular}%
}
\caption{Results of stigma detection task on our corpus. \textit{P}, \textit{R}, \textit{F1}, and \textit{Acc} stand for macro precision, macro recall, macro F1, and accuracy respectively. The best performance is colored in \colorbox[rgb]{1.0, 0.875, 0.894}{pink}. See Appendix \ref{app:full_results} for a detailed breakdown of model performance on responses to each of the seven interview questions.}
\label{tab:performance}
\vspace{-4.5pt} 
\end{table*}

\subsection{Experimental Results}

Our experimental results in Table \ref{tab:performance} show that detecting mental-health stigma remains challenging, as we expected. 
We observed that performance generally scales with model size, with GPT-4o and Llama-3.3-70B performing best, followed by Mixtral 8×7B and Mistral Nemo, while Llama-3.1-8B lagged behind. 
We also noted consistent improvements across the models when we enriched the prompts with more detailed information, with GPT-4o's F1 score increasing from 0.456 to 0.576 when a single example was added to the prompt. 
The gains were even more substantial when we provided the full codebook in the prompt -- which included label definitions, representative keywords per label, human annotator-derived rules/specifications, along with the same example as in the one-shot prompt -- as evidenced by Llama-3.3-70B's F1 score increasing from 0.545 to 0.752.

In contrast to this trend, however, Mixtral 8×7B achieved a higher accuracy (0.608) than GPT-4o (0.578) in the zero-shot setting, which can probably be explained by the highly unbalanced data distribution in our corpus -- 42 \textit{Stigmatizing (pity)} instances versus 2,232 \textit{Non-stigmatizing} ones -- where predicting the majority class inflates accuracy. 
Interestingly, when given prompts with detailed guidance, the models showed higher recall at the expense of precision, as exemplified by GPT-4o achieving a recall of 0.801 but a precision of 0.742 in the full-codebook setting, indicating better detection of stigmatizing language but more false positives.

To sum up, these results underscore the need for human-generated guidelines and confirm that reliance on neural models alone remains insufficient to capture an array of subtleties in mental-health stigma, substantiating the development of our interview corpus as a much-needed benchmark.

\subsection{Challenges in Detecting Mental-Health Stigma}

To further understand the challenges of stigma detection, we empirically and qualitatively investigated all 137 misclassified interview snippets (out of 829 total) from GPT-4o's predictions using full-codebook prompting -- one of our best-performing experimental setups -- and uncovered a set of deeply embedded, socially normalized stigmatizations that the model struggled to identify correctly.

\textit{Linguistically}, we observed several recurring patterns in these misclassified utterances: 1) the use of \textbf{distancing language} \cite{distancing_language_nook_2017}, where speakers employ third-party perspectives to disguise their personal views, for example, "\textit{Neighbors may find it hard to understand Avery's outbursts and strange behavior if they do not know them very well.}" (P388), 2) the dismissive \textbf{misappropriation of psychiatric terminology} \cite{misappropriation_lilienfeld_2015}, such as describing people with mental illness as "\textit{paranoid}" (P28) without proper context or medical basis, and 3) \textbf{coercive phrasing} in advice-giving, where speakers use terms like "\textit{definitely need}" to impose decisions on people with mental illness rather than offering "\textit{suggestions}" that respect their autonomy and agency.

\textit{Semantically}, we identified certain microaggressions that models often fail to detect -- 1) \textbf{differential support} \cite{special_care_iseselo_2016} emerges when participants display excessive caution toward people with mental illness, subconsciously positioning them as inferior or in need of special treatment, with participants expressing a need to "\textit{be more mindful}" or "\textit{humble oneself}" during interactions (P510), 2) \textbf{patronization/paternalism} \cite{patronization_douglas_2011} surfaces in condescending and demeaning attitudes where speakers position themselves as authorities who can "\textit{teach}" people with mental illness proper ways to live (P584), and 3) \textbf{trivialization/minimization} \cite{underestimate_hopkins_2014} appears when speakers downplay the legitimate challenges inflicted by mental-health conditions. 
Representative quotes are provided in Appendix \ref{app:error}.

\section{Conclusion}

As an initial effort, our expert-annotated interview corpus, informed by the socio-conceptual framework \cite{attribution_model_corrigan_2003} and collected through human-chatbot conversations, can serve as an infrastructure to facilitate detect and finely classify mental-health stigma.
By further documenting the socio-cultural context, such as the gender of interviewees, this corpus allows for data lineage tracing and partly prevents mis/underrepresentation of certain social groups. 
In addition, the empirical results illustrate existing challenges for state-of-the-art LLMs in decoding those seemingly well-intentioned stigmatizing expressions, suggest areas for improvement in computational approaches, and thus contribute to both computational social science and the NLP community.

Our keystone dataset also provides important implications for future research. 
First, it can benchmark the extent to which \textbf{neural models internalize and perpetuate stigma} by having them role-play interviewees and comparing their generated responses with real-world, human-provided ones.
Second, it captures \textbf{conversation dynamics} \cite{conversation_dynamic_hua_2024} of how humans and chatbots navigate stigma-related discussions, which can inform the development of empathetic conversational agents.
Third, it opens new avenues for research on computationally \textbf{neutralizing, reducing, and counteracting} \cite{counteract_podolak_2024} stigma, which could potentially be generalized to other psychological constructs \cite{psych_construct_meng_2025} where attribution models are applicable (e.g., LGBTQ+ stigmatization \cite{lgbtq_stigma_haider_2008}).
Finally, it offers insights into \textbf{causal-reasoning} patterns \cite{causal_graph_meng_2025} and \textbf{moral judgments/values} \cite{moral_haidt_2007} underlying stigma, which could guide the design of personalized interventions and stigma-reduction campaigns.

\section*{Limitations}

We recognize that our paper warrants discussion of several limitations.

\paragraph{\textcolor{darkred}{\textit{Multi-Perspective Annotation.}}}

Our dataset release includes consolidated labels from two annotators, yet we acknowledged that mental-health stigma represents a deeply subjective phenomenon shaped by socio-psychological and cultural factors \cite{culture_difference_krendl_2020}. 
The interpretation and annotation of stigmatizing content inherently vary according to the annotators' backgrounds, experiences, and temporal contexts \cite{multi_label_rottger_2022, multi_label_paun_2021}. 
Recent work has emphasized the critical role of annotators' demographics and beliefs in shaping data labels for sensitive social tasks \cite{multi_label_sap_2022, multi_label_wan_2023, multi_label_giorgi_2024}. 
The perspectivist paradigm further suggests capturing this natural variation rather than enforcing consensus \cite{multi_label_fleisig_2024, multi_label_prabhakaran_2021}. 
An important next step would be to unlock richer perspectives by releasing version 2.0 of our corpus with \textbf{annotator-level labels} that preserve individual viewpoints and disagreements.

\paragraph{\textcolor{darkred}{\textit{Cultural Sensitivity.}}}

The current dataset is largely drawn from Western, English-speaking sources, yet mental-health stigma remains deeply interrelated with and tied to cultural values, belief systems, and social norms. 
Research has demonstrated the unique challenges of analyzing stigma and stereotypes in different social groups and languages \cite{multi_culture_fort_2024}.
It can be predicted that internalized prejudice, help-seeking barriers, and social-exclusion dynamics will differ markedly across multinational, multilingual, and multicultural settings. 
Accordingly, a promising direction is to expand our corpus to version 3.0 by collecting and annotating stigma-related discussions from Eastern regions, allowing for cross-cultural analysis of how social stigma is unfolded, negotiated, and expressed.

\paragraph{\textcolor{darkred}{\textit{Intersectionality.}}}

Our corpus does not explicitly account for intersectionality in mental-health stigma, which often interacts with other forms of stereotyping and prejudice, such as racism, ageism, and misogyny \cite{intersectionality_lin_2022}. 
The interplay of these intersectional biases proves difficult to operationalize, as they manifest in complex, jointly reinforcing patterns that compound and affect each other. 
A natural extension of our work would be to develop finer-grained annotation schemes that can capture these intersectional dynamics while maintaining analytical clarity.

\paragraph{\textcolor{darkred}{\textit{Prompt Robustness.}}}
The prompt design and few-shot example selection in our experiments, while functional, could benefit from a more systematic evaluation. 
The specific choice of phrasing and exemplar selection may influence model behavior in ways that we have not thoroughly tested. 
In addition, the order of stigma attributions remains the same in both the one-shot and full-codebook prompt settings, which may introduce primacy and/or recency biases \cite{prompt_sensitivity_lu_2022}.
Fruitful avenues for future research include conducting ablation studies through structured variation of prompt components, example counts (e.g., five-shot), and linguistic patterns.

\paragraph{\textcolor{darkred}{\textit{Pre-existing Model Biases.}}}

Previous studies have shown that LLMs and pre-trained language models (PLMs) harbor and inadvertently reinforce societal biases and/or stereotypes, including those related to mental health \cite{llm_bias_mina_2024, intersectionality_lin_2022}. Therefore, our experimental results may be unavoidably influenced by these flawed pre-existing biases. We should remain alert to the possibility that LLM/PLM biases affect our findings, and an important next step would be to evaluate how these biases impact model performance.

\section*{Ethics and Broad Impact}

Our corpus creation and annotation processes followed rigorous ethical protocols with full Institutional Review Board (IRB) approval (NUS-IRB-2024-391), and we implemented comprehensive consent procedures during data collection in which participants received detailed information about data storage, use, and release policies. 
To protect privacy, our published dataset includes only responses to the interview and follow-up questions, with all personal identifiers removed. 

We established ongoing monitoring mechanisms for our research assistants who served as annotators -- the main researcher conducted regular check-ins to assess any potential impact on their mental well-being from exposure to stigmatizing content, with their feedback documented in Appendix \ref{app:feedback}.

We recognize the potential risks associated with releasing this mental-health stigma corpus. 
The primary concern is that models trained on this data may inadvertently amplify existing biases and stereotypes against people with mental illness.
We have implemented ethical guardrails, including documentation of dataset limitations, access request forms requiring researchers to specify intended uses, and feedback channels for reporting ethical concerns, and strongly advocate for the responsible use of this dataset through regular audits, careful deployment considerations, and appropriate research applications.

\section*{Acknowledgments}

We thank the anonymous reviewers for their feedback on this work. 
This work was partially funded by the National University of Singapore CSSH (24-1774-A0002), the National University of Singapore HSS Seed Fund CR (2024 24-1191-A0001), and the Ministry of Education Tier 1 (24-1317-A0001).

\bibliography{ref}

\appendix

\section{Discussion of Problematic Language Definition}
\label{app:term}

We would like to clarify and distinguish between several concepts commonly studied by NLP scholars \cite{classify_term_hate_fortuna_2020, classify_term_hate_pachinger_2023} and the terms we use throughout this paper. 
\textbf{Abusive language} refers to content that ascribes a negatively judged social identity to individuals, marking them as shameful or morally objectionable representatives of a marginalized group \cite{abusive_wiegand_2019}. 
\textbf{Offensive language} encompasses insults, profanity, and targeted attacks that may harm disadvantaged groups \cite{offensive_davidson_2017}. 
\textbf{Hate speech} specifically involves expressing hatred or intending to degrade members of protected groups based on characteristics like race, gender, or disability \cite{hate_speech_waseem_2016}. 
\textbf{Toxic language} more broadly covers disrespectful or inappropriate content that is likely to cause people to leave discussions \cite{toxic_palmer_2021}.

On the other hand, \textbf{social stigma} represents society's collective negative attitudes, prejudices, and discriminatory practices that devalue and exclude certain individuals or groups \cite{stigma_spoiled_identity_goffman_1964}. 
Unlike these forms of problematic language, which can be directly observed, inferred, and/or detected, it exists as a latent psychological construct with deep underpinnings rooted in classic theories \cite{attribution_model_corrigan_2003, self_stigma_link_1989}, and \textbf{mental-health stigma} specifically targets people with mental illness. 
This means that it needs to be operationalized and measured in terms of components -- cognitive judgments (i.e., stereotypes), emotional responses (i.e., prejudice), and behavioral responses (i.e., discrimination) \cite{stigma_spoiled_identity_goffman_1964} -- that language might be able to capture.

\begin{figure*}[ht]
    \centering
    \includegraphics[width=\linewidth]{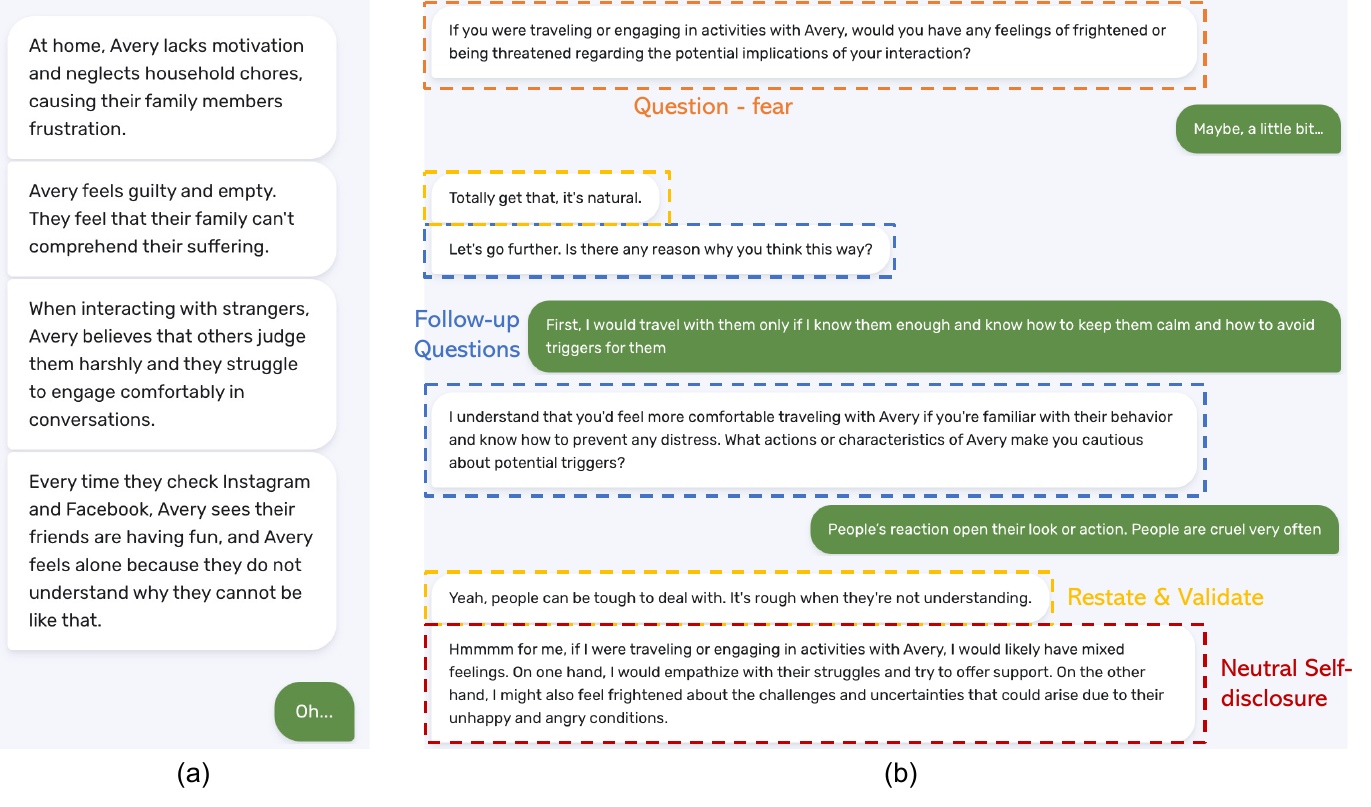}
    \caption{Chatbot interface for (a) vignette delivery and (b) an example interview question, showing a multi-turn conversation about the attribution of \textit{fear} between the chatbot and participant P189.}
    \label{fig:interface}
\end{figure*}

\section{More Details about Data Collection: Chatbot-based Interview}
\label{app:collection}

\subsection{Vignettes}

The clinical version appears below:

\begin{quote}
    \textit{Avery is employed by a company, and in their spare time, they are dedicated to lifelong learning, doing extensive reading and writing. However, Avery has been diagnosed with depression recently. It has become challenging for them to concentrate, resulting in a decline in work performance and learning outcomes. Interests that once brought them joy have waned. Avery has distanced themself from friends, becoming easily irritated in social interactions with colleagues and feeling judged by others. Avery lives with family and cannot do much, especially household chores. Social media intensifies their feelings of loneliness and anger, leading to frustration about the source of the anger.}
\end{quote}

The narrative version delivered to participants can be found in Table \ref{tab:sample}.

\subsection{Chatbot Design Elements}

To facilitate participants' self-disclosure, we designed our chatbot with three key strategies. 
First, we humanized the interaction and added \textbf{anthropomorphic} elements \cite{anthropomorphism_abercrombie_2021} by naming it \textit{Nova}. 
Second, Nova employed \textbf{active listening} \cite{active_strategies_zheng_2023}, where it restated and validated the user's point of view and expressed its understanding. 
And third, it used \textbf{neutral self-disclosure} \cite{disclosure_lee_2022}, sharing both positive and negative perspectives without shifting their attitudes to encourage reciprocal openness.

On the other hand, to alleviate potential emotional distress, the vignette presentation was punctuated with intermittent prompts soliciting brief responses from participants \cite{chatbot_aq27_practice_lee_2023, stigma_reduce_cui_2024}, such as asking what advice they might offer to Avery. 
In addition, we strategically positioned a \textit{mid-session break} between the seven interview questions, where the conversation shifted to lighthearted dialogue about hobbies, serving to re-engage participants and provide emotional decompression.

\subsection{Implementation}

Our chatbot combines \textbf{pre-written scripts} with \textbf{outputs from AI models} to create a natural conversational flow. 
The core elements -- small-talk questions, vignette delivery, interview questions ($\mathcal{IQ}s$), and the chatbot's neutral self-disclosure -- were pre-scripted and hard-coded. 
The AI model generates all other elements of the interview, including crafting follow-up questions ($\mathcal{FQ}s$) and formulating responses to participant input. 
For example, when generating $\mathcal{FQ}_2$ for the \textit{anger} attribution, we used the following prompt:

\begin{quote}
\texttt{- Chatbot: Thank you for sharing! Let's imagine you were one of Avery's friends at the party or other social occasions. Would you feel anger towards them due to their actions, for example, losing their temper and yelling at someone?}

\texttt{- User: [Answer 1].}

\texttt{- Chatbot: May I kindly ask why?}

\texttt{- User: [Answer 2].}

\texttt{Given this conversation context where the chatbot asks an interview question and the user provides a response, generate a follow-up that: 1) demonstrates understanding by restating the user's perspective, 2) asks a single ``how" or ``what" question about Avery to encourage detailed self-disclosure, and 3) explores which specific plot elements or actions led to their thinking. The response should be conversational, under 30 words, use first-person pronouns (``I," ``me"), and refer to Avery as ``them."}
\end{quote}

We implemented the model using \texttt{GPT-4-1106-preview} \cite{gpt4_achiam_2023}, with a maximum token limit of 100 and temperature set to 0.2 for consistent output.
The chatbot interface, shown in Figure \ref{fig:interface}, was integrated into the Qualtrics survey, and we ensured that the concurrency rate remained below 50.


\subsection{An Example of A Complete Interview Script}

Table \ref{tab:sample} shows an example of a complete interview flow. 
Our released corpus contains only the seven question-answer segments ($\mathcal{Q}1$-$\mathcal{Q}7$) for ease of use.

\section{Interview Participant Demographics}
\label{app:demographic}

Table \ref{tab:participant} presents the self-reported demographic and geographic characteristics of 555 out of 684 participants (81.10\%) who voluntarily shared this information and served as our primary data source.

\begin{table*}[t]
  \centering
  \small

    \begin{tabular}{llc}
    \toprule
    \multicolumn{2}{c}{} & \multicolumn{1}{p{11em}}{\textbf{ALL (N=555) n (\%)}} \\
    \midrule
    \multirow{3}[2]{*}{\textbf{Gender}} & Female & 305 (54.95) \\
    & Male  & 249 (44.86) \\
          & Prefer not to say & 1 (0.18) \\
    \midrule
    \multirow{6}[2]{*}{\textbf{Age}} & 21-24 & 61 (10.99) \\
          & 25-34 & 160 (28.83) \\
          & 35-44 & 109 (19.64) \\
          & 45-54 & 75 (13.51) \\
          & 55-64 & 67 (12.07) \\
          & 65+   & 83 (14.95) \\

    \midrule
    \multirow{6}[2]{*}{\textbf{Ethnicity}} & White & 356 (64.14) \\
          & Black or African American & 133 (23.96) \\
          & Asian & 36 (6.49) \\
          & Mixed & 19 (3.42) \\
          & American Indian or Alaska Native & 1 (1.80) \\
          & Other (Hispanic, Chicano, etc.) & 10 (1.8) \\
    \midrule
    \multirow{7}[2]{*}{\textbf{Country}} & United States & 187 (33.69) \\
          & United Kingdom & 146 (26.31) \\
          & South Africa & 98 (17.66) \\
          & Canada & 43 (7.75) \\
          & Australia & 28 (5.05) \\
          & Ireland & 18 (3.24) \\
          & Other (13 countries) & 35 (6.31) \\
    \midrule
    \multirow{9}[2]{*}{\textbf{Education}} & Less than primary & 1 (0.18) \\
          & Primary & 3 (0.54) \\
          & Some secondary & 4 (0.72) \\
          & Secondary & 83 (14.95) \\
          & Vocational or similar & 62 (11.17) \\
          & Some University but no degree & 94 (16.94) \\
          & University - Bachelor's degree & 197 (35.50) \\
        & \multicolumn{1}{p{17.915em}}{Graduate or professional degree (MA, MS, MBA, PhD, law degree, medical degree, etc.)} & \multirow{2}{*}{109 (19.64)} \\
          & Prefer not to say & 2 (0.36) \\
            \midrule
    \multirow{3}[2]{*}{\textbf{Mental-illness Experience}} & Yes   & 320 (57.66) \\
          & No    & 133 (23.96) \\
          & Maybe & 102 (18.38) \\
    \bottomrule
    \end{tabular}%
      \caption{Participant Demographics. \textbf{Mental-illness experience} refers to whether participants had immediate family members or close friends who experienced mental illness.}
  \label{tab:participant}%
\end{table*}%

\section{More Details about Data Annotation}
\label{app:annotation}

\subsection{Description of Stigma Attributions}
\label{app:attributions}

Our annotation scheme categorizes texts into either non-stigmatizing attitudes or one of seven stigma attributions that capture different facets of mental-health stigmatization:

\begin{itemize}
\vspace{-3pt}
\item \textbf{Stigmatizing (Responsibility)}: Believing people have control over and are responsible for their mental illness and related symptoms.\vspace{-5pt}
\item \textbf{Stigmatizing (Social Distance)}: Staying away from people with mental illness.\vspace{-5pt}
\item \textbf{Stigmatizing (Anger)}: Expressing irritation or annoyance toward people with mental illness.\vspace{-5pt}
\item \textbf{Stigmatizing (Helping)}: Withholding support toward people with mental illness.\vspace{-5pt}
\item \textbf{Stigmatizing (Pity)}: Being unsympathetic toward people with mental illness.\vspace{-5pt}
\item \textbf{Stigmatizing (Coercive Segregation)}: Forcing institutionalization and mandatory treatment on people with mental illness.\vspace{-5pt}
\item \textbf{Stigmatizing (Fear)}: Perceiving people with mental illness as dangerous, unpredictable, and unsafe to be around.\vspace{-5pt}
\item \textbf{Non-stigmatizing}: Showing understanding, empathy, and support toward people with mental illness, recognizing mental-health challenges as complex medical conditions influenced by multiple factors.
\end{itemize}

\subsection{Annotation Platform}

\begin{figure*}[ht]
    \centering
    \includegraphics[width=1\linewidth]{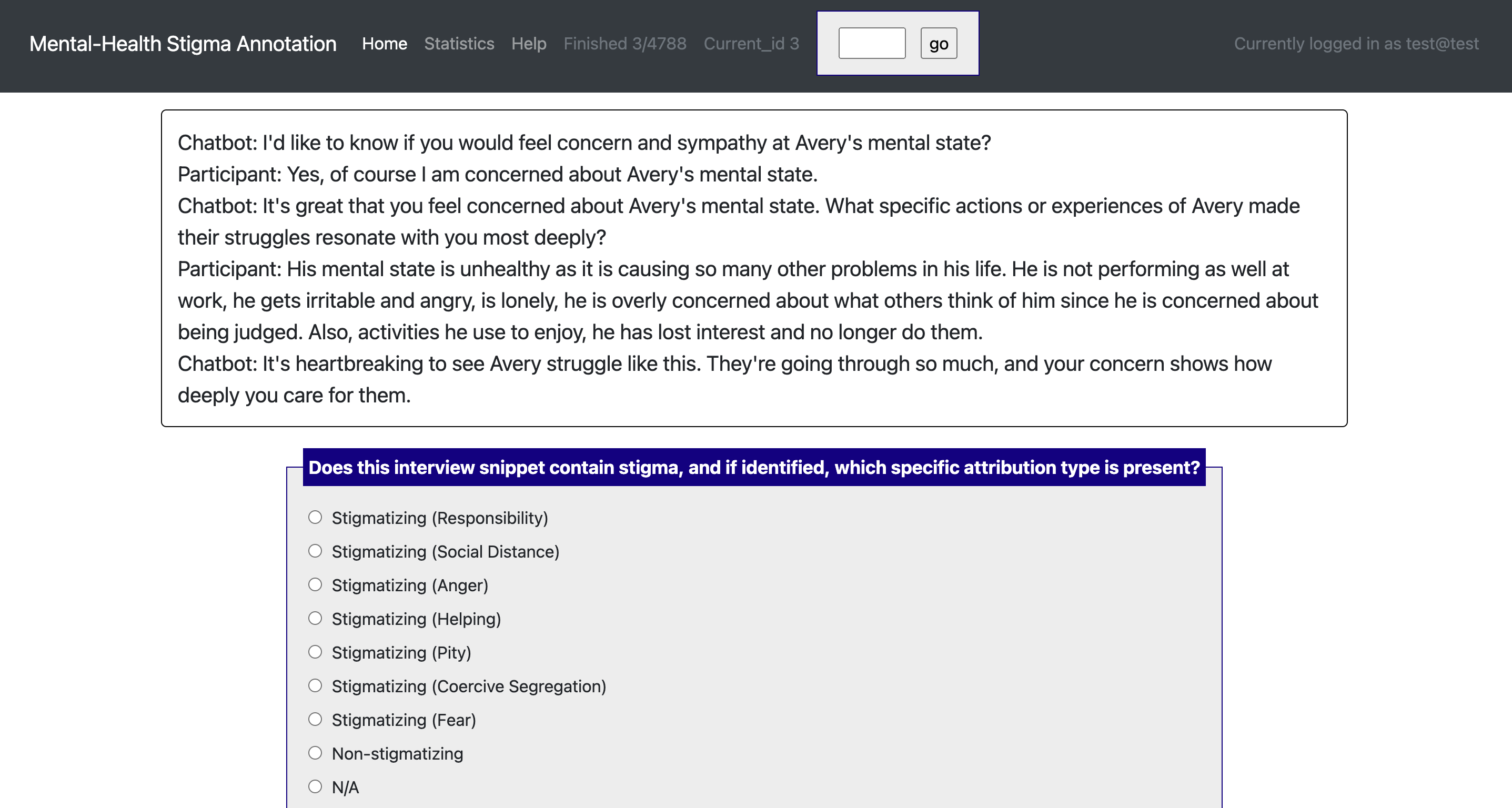}
    \caption{Screenshot of the annotation platform interface.}
    \label{fig:annotation}
\end{figure*}

Figure \ref{fig:annotation} shows the screenshot of the annotation platform interface.

\subsection{Annotation Instructions}
\label{app:codebook}

\begin{figure*}[ht]
    \centering
    \includegraphics[width=1\linewidth]{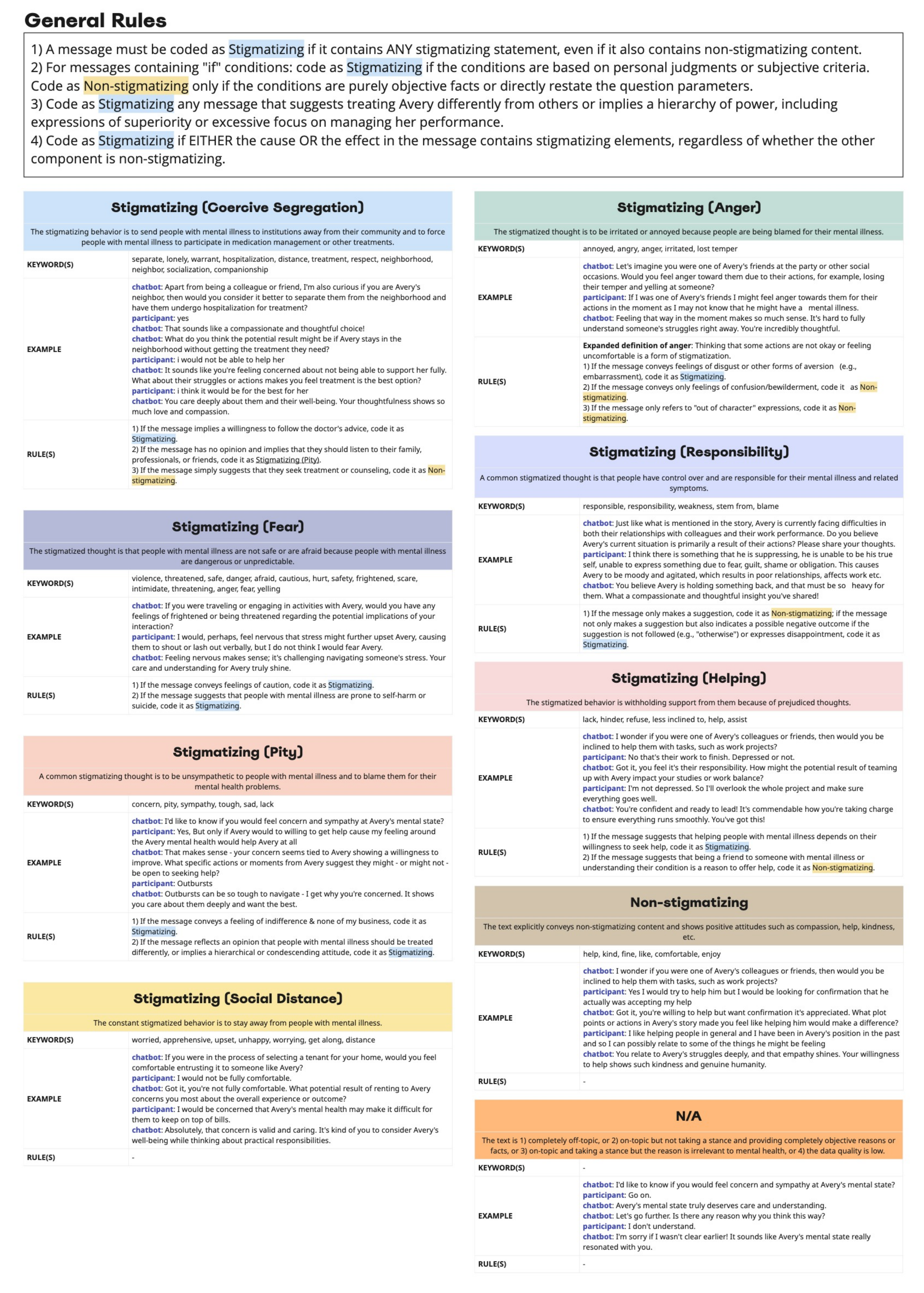}
    \caption{Annotation instructions for human annotators.}
    \label{fig:codebook}
\end{figure*}

It should be noted that our annotation instructions were iteratively refined through active collaboration with both annotators, who provided valuable input and suggestions based on their hands-on coding experience, rather than being passive recipients of predetermined guidelines. 
See Figure \ref{fig:codebook} for our detailed annotation instructions shown to human annotators.

\subsection{Agreement Matrix between Human Annotators}
\label{app:matrix}

\begin{figure*}[ht]
    \centering
    \includegraphics[width=0.74\linewidth]{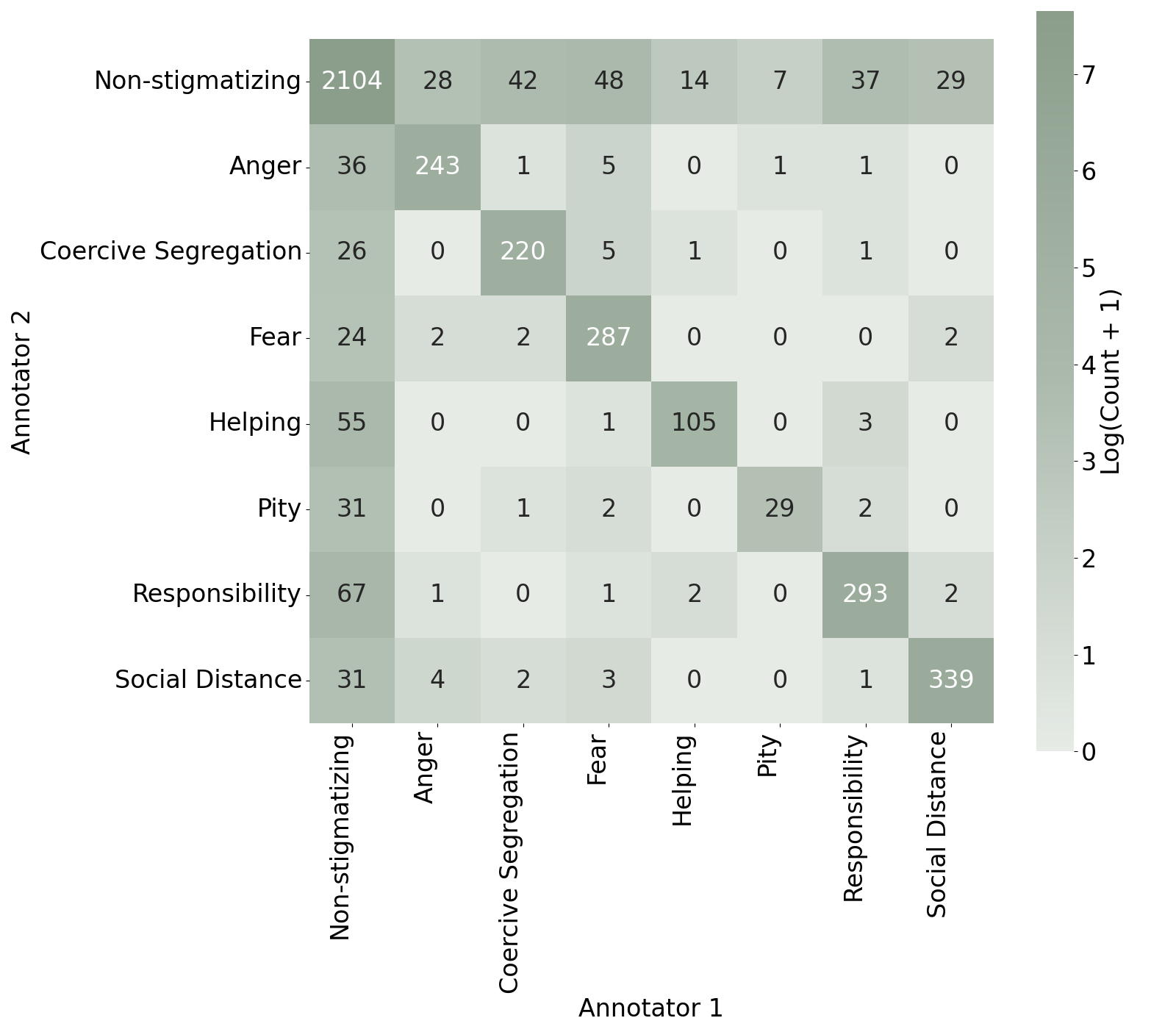}
    \caption{Heatmap showing the agreement between two human annotators.}
    \label{fig:heatmap_human}
\end{figure*}

Figure \ref{fig:heatmap_human} presents the heatmap showing the agreement between the two human annotators. 
The matrix reveals relatively low confusion between different stigma attributions, while most disagreement occurs when one annotator labels a response as non-stigmatizing and the other identifies it as containing a specific type of stigma.

\subsection{Annotator Feedback}
\label{app:feedback}

\subsubsection{Reflexive Impact on Annotators' Perspectives}

The annotation experience fostered reflexive awareness among annotators, though their responses to stigmatized content varied. 
One annotator developed a heightened sensitivity to implicit discrimination in everyday discourse, becoming more conscious of their own language choices and gaining deeper empathy for stigmatized groups. 
The other annotator maintained their established viewpoints, only occasionally engaging in self-reflection about their attitudes during the annotation process.

\subsubsection{Risks of Over-interpretation}

We noticed that the long annotation timespan, combined with the subtle nature of stigma, created challenges in maintaining consistent judgment standards. 
Interestingly, one annotator found themselves detecting multiple stigma attributions in single interview snippets and noticing forms of stigma not captured by the existing scheme, initially worrying about over-interpretation and over-analysis. 
However, these observations presumably revealed the complexity of stigmatized language in real conversations rather than analytical overreach. 
Their insights suggest valuable opportunities for exploring how different stigma attributions co-exist and intertwine, and for better conceptualizing more implicit forms of stigmatization.

\subsubsection{Training and Knowledge Requirements}

Despite initial unfamiliarity with the topic, annotators reported that they found the task accessible through proper training and communication with co-workers and specialists. 
The codebook evolved through multiple iterations, gained sharper definitions, and offered better guidance for ambiguous cases.
Notably, one annotator emphasized that rather than requiring extensive prior knowledge, the main challenge lay in accurately understanding and applying the annotation rules, especially in borderline cases.

\section{More Corpus Statistics}
\label{app:more_stat}

\begin{figure*}[t]
    \centering
    \begin{subfigure}[b]{0.48\textwidth}
        \includegraphics[width=\textwidth]{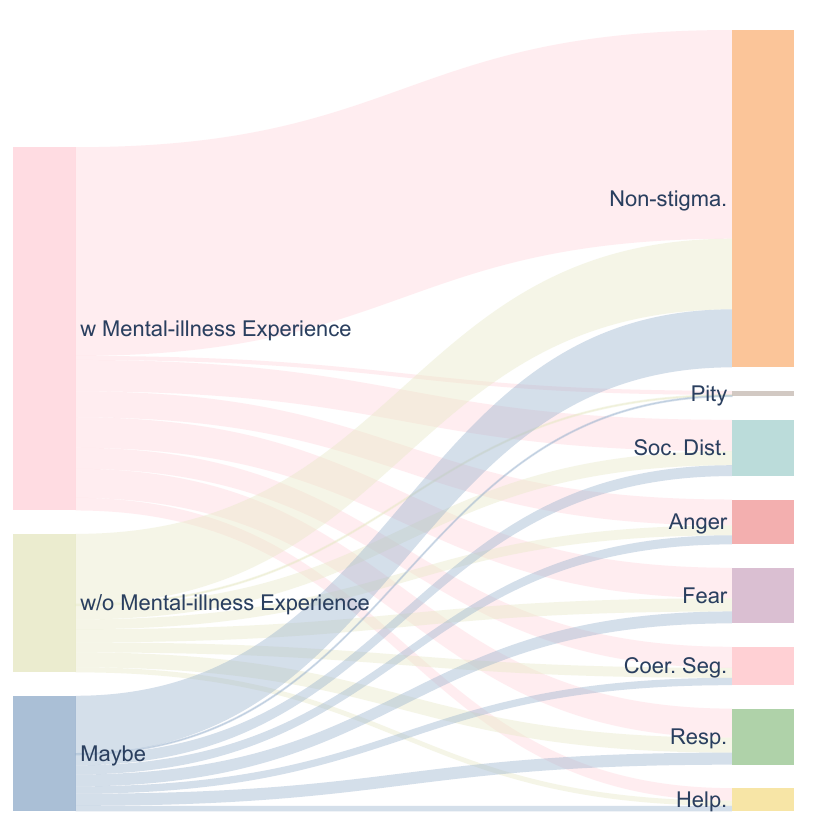}
    \end{subfigure}
    \hfill
    \begin{subfigure}[b]{0.51\textwidth}
        \includegraphics[width=\textwidth]{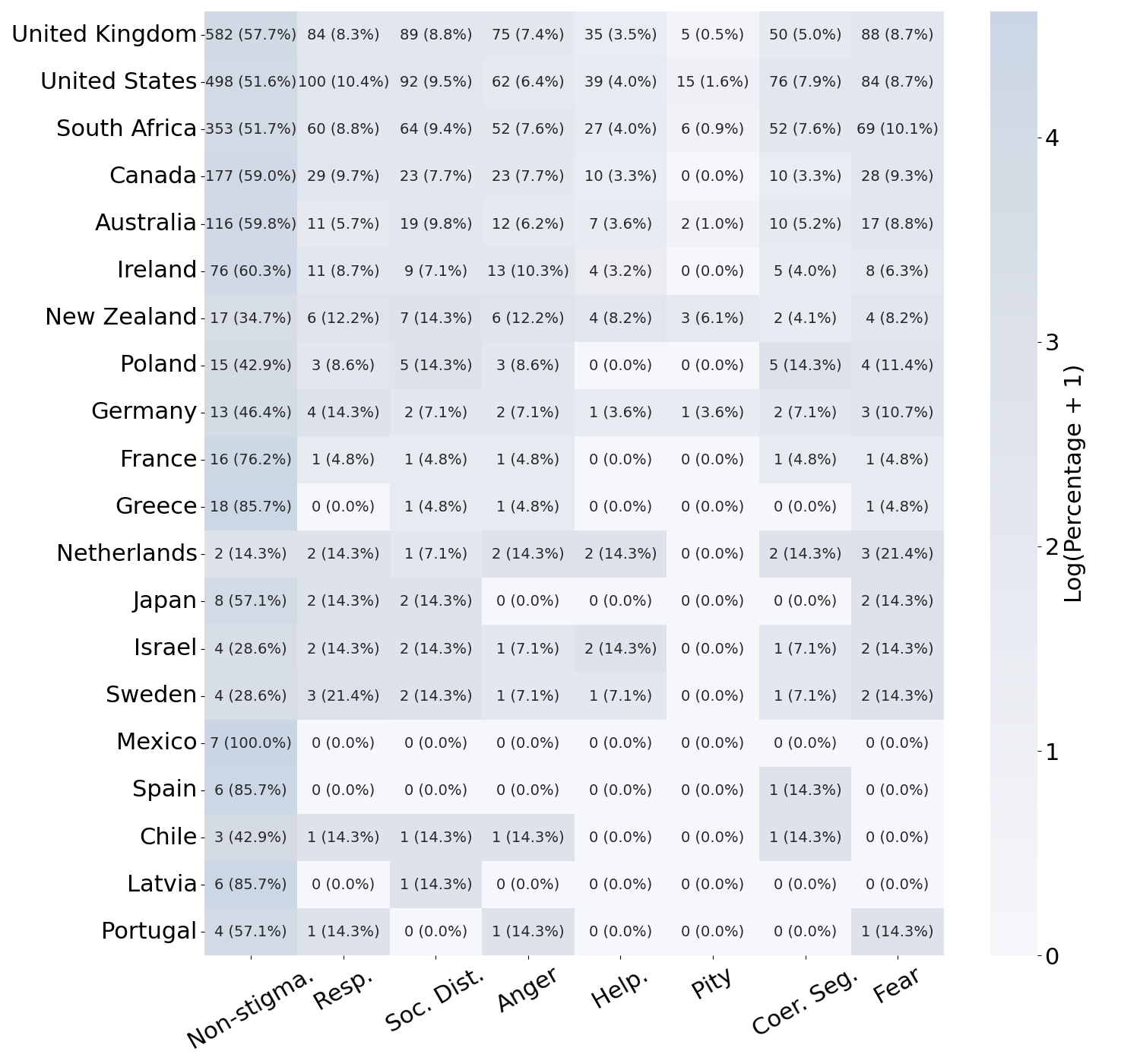}
    \end{subfigure}
\caption{More association between mental-health stigma and sociocultural factors in our corpus: mental-illness experience (left) and country of residence (right).}
    \label{fig:socio}
\end{figure*}

\begin{figure*}[t]
    \centering
     \includegraphics[width=\textwidth]{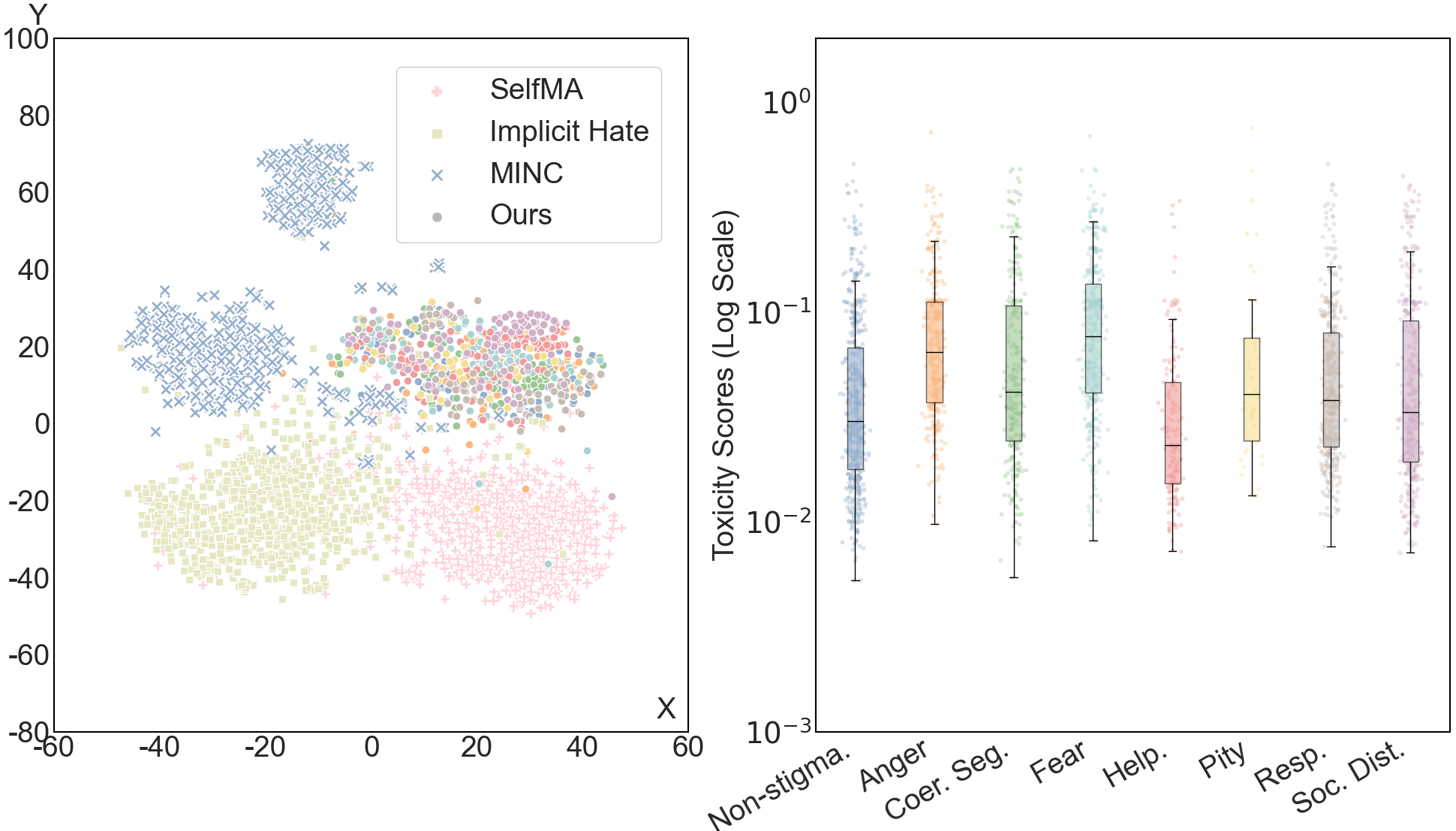}
\caption{Word-embedding comparison of 1,000 random samples each from \textsc{SelfMA} \cite{microaggression_dataset_breitfeller_2019}, \textsc{Implicit Hate Speech} \cite{hate_speech_dataset_elSherief_2021}, \textsc{MINC} \cite{mental_health_stigma_corpus_mitkov_2023} and our corpus (left); and toxicity-score distributions across different labels in our corpus (right).}
    \label{fig:more_stats}
\end{figure*}

For additional statistical analysis of our proposed corpus, see Figure \ref{fig:socio} and \ref{fig:more_stats}.

\subsection{Stigma-emotion Correlations}
\label{app:emotion}

Following \citet{emotion_method_li_2024}, we used the pre-trained BERT-based emotion classifier\footnote{\url{https://huggingface.co/bhadresh-savani/bert-base-uncased-emotion}} to detect six emotion categories (\textit{joy}, \textit{love}, \textit{surprise}, \textit{anger}, \textit{fear}, and \textit{sadness}) as established in \citet{emotion_foundation_shaver_1987}. 
We performed OLS regression analysis to quantify associations between stigma attributions and emotion categories.

Results shown in Table \ref{tab:emotion_correlation} show that \textbf{Non-stigmatizing} responses correlate positively with \textit{joy} ($b=0.1731$, $p<0.001$) and \textit{love} ($b=0.0387$, $p<0.01$), but negatively with \textit{anger} ($b=-0.1116$, $p<0.001$). 
In contrast, \textbf{Stigmatizing (anger)} responses correlate negatively with \textit{joy} ($b=-0.2471$, $p<0.01$) and positively with \textit{anger} ($b=0.4483$, $p<0.001$). 
\textbf{Stigmatizing (fear)} responses also correlate negatively with \textit{joy} ($b=-0.2460$, $p<0.01$), but positively with both \textit{fear} ($b=0.2699$, $p<0.001$) and \textit{sadness} ($b=0.1743$, $p<0.01$).
Together, these correlations substantiate the value of our corpus for examining the psycho-emotional aspects of mental-health stigma.

\begin{table*}[tbp]
\small
\renewcommand{\arraystretch}{1.3}
\centering
\begin{tabular}{p{5cm}cccccc}
\toprule
& \textbf{Joy} & \textbf{Love} & \textbf{Surprise} & \textbf{Anger} & \textbf{Fear} & \textbf{Sadness} \\
\midrule
\multirow{2}{5cm}{Non-stigmatizing} & 0.1731*** & 0.0387** & -0.0063 & -0.1116*** & -0.0279 & 0.0451 \\
& (0.000) & (0.001) & (0.072) & (0.000) & (0.214) & (0.065) \\[0.5ex]
\multirow{2}{5cm}{Stigmatizing (anger)} & -0.2471** & -0.0102 & -0.0195* & 0.4483*** & -0.0999 & 0.0394 \\
& (0.002) & (0.750) & (0.034) & (0.000) & (0.091) & (0.540) \\[0.5ex]
\multirow{2}{5cm}{Stigmatizing (coercive segregation)} & 0.3971*** & -0.0044 & 0.0032 & -0.0672 & -0.0768 & -0.1408* \\
& (0.000) & (0.888) & (0.723) & (0.328) & (0.181) & (0.024) \\[0.5ex]
\multirow{2}{5cm}{Stigmatizing (fear)} & -0.2460** & -0.0134 & -0.0094 & -0.0643 & 0.2699*** & 0.1743** \\
& (0.001) & (0.660) & (0.285) & (0.340) & (0.000) & (0.005) \\[0.5ex]
\multirow{2}{5cm}{Stigmatizing (helping)} & -0.1494 & 0.0451 & -0.0173 & 0.0687 & 0.1401 & 0.0239 \\
& (0.151) & (0.289) & (0.159) & (0.466) & (0.076) & (0.780) \\[0.5ex]
\multirow{2}{5cm}{Stigmatizing (pity)} & 0.1579 & -0.0085 & 0.0740*** & 0.0122 & -0.0799 & -0.0446 \\
& (0.302) & (0.892) & (0.000) & (0.930) & (0.490) & (0.723) \\[0.5ex]
\multirow{2}{5cm}{Stigmatizing (responsibility)} & 0.1936** & 0.0001 & -0.0059 & -0.0826 & 0.0150 & -0.0092 \\
& (0.005) & (0.996) & (0.466) & (0.183) & (0.772) & (0.870) \\[0.5ex]
\multirow{2}{5cm}{Stigmatizing (social distance)} & -0.0519 & -0.0235 & -0.0091 & 0.0806 & 0.0388 & 0.0762 \\
& (0.479) & (0.431) & (0.292) & (0.225) & (0.485) & (0.206) \\
\bottomrule
\end{tabular}
\caption{Correlation coefficients between stigma attributions and emotion categories. $p$-values are shown in parentheses. Significance levels are denoted as ***: $p<0.001$, **: $p<0.01$, *: $p<0.05$.}
\label{tab:emotion_correlation}
\end{table*}

\subsection{Response-quality Pattern}

\begin{table*}[h]
\small
\renewcommand{\arraystretch}{1.15}
\centering
\begin{tabular}{lrrrr}
\toprule
\textbf{Response Type} & \textbf{\# Responses} & \textbf{\# Participants} & \textbf{Avg Length} & \textbf{\# Responses} \\
& \textbf{(\% total)} & \textbf{(\% total)} & \textbf{(chars)} & \textbf{still < 150 chars after $\mathcal{FQ}s$} \\
\midrule
$\mathcal{IQ}$ only (no $\mathcal{FQ}s$ needed) & 609 (14.71\%) & 250 (36.55\%) & 219.54 & 0 \\
$\mathcal{IQ}$ + one $\mathcal{FQ}$ & 2,459 (59.38\%) & 639 (93.42\%) & 167.71 & 1,138 \\
$\mathcal{IQ}$ + two $\mathcal{FQ}s$ & 1,073 (25.91\%) & 392 (57.31\%) & 116.99 & 805 \\
\bottomrule
\end{tabular}
\caption{Interview response statistics by number of $\mathcal{FQ}s$ asked. We report the count and percentage of interview snippets in which zero, one, or two $\mathcal{FQ}s$ were asked, the number of participants who contributed at least one snippet to each group (noting that one participant can contribute up to seven interview snippets and may therefore be counted in more than one group), the average total character length of participants' responses to the $\mathcal{IQ}$ and any $\mathcal{FQ}s$, and the number of snippets with participant responses that remained below our 150-character threshold even after all $\mathcal{FQ}s$ were asked.}
\label{tab:response_patterns}
\end{table*}

As shown in Table \ref{tab:response_patterns}, 14.71\% of the responses (from 36.55\% of the participants) naturally exceeded the 150-character threshold without requiring follow-up questions, averaging 219.54 characters ($SD=79.37$). 
In contrast, responses requiring one follow-up question (59.38\%) averaged 167.71 characters ($SD=78.87$), while those requiring two follow-up questions (25.91\%) averaged only 116.99 characters ($SD=62.30$). 
After follow-up prompts, 46.28\% of responses with one $\mathcal{FQ}$ and 75.02\% with two $\mathcal{FQ}s$ were still below our 150-character threshold.

Together with our familiarity with the data, these results allow us to qualitatively infer that responses exceeding 150 characters without requiring $\mathcal{FQ}s$ tended to be of higher quality, with more coherent language, deeper thoughts, and richer, more interesting content.
For example, we sometimes observed participants incorporating factors such as \textbf{past experience} and \textbf{personality} into their reasoning pertaining to mental health.

\begin{table*}[t]

\begin{tabular}{@{}p{\textwidth}@{}}
    \toprule
    \textcolor{darkred}{\textbf{Small Talk}}\\
    \midrule
    \textbf{Chatbot (Nova)}: Hi, I'm Nova. Nice to meet you! Could you please tell me your nickname? (Please just input your nickname!) \\
    \textbf{Participant}: \textsc{user\_name} \\
    \textbf{Chatbot (Nova)}: Welcome \textsc{user\_name}! I'm overjoyed to have the opportunity to work with you. \\
    \textbf{Chatbot (Nova)}: How was your mood today? \\
    \textbf{Participant}: just ok \\
    \textbf{Chatbot (Nova)}: Hope things get better for you soon. \\
    \textbf{Chatbot (Nova)}: What is your favorite activity after finishing work or school? \\
    \textbf{Participant}: Go out and exercise. \\
    \textbf{Chatbot (Nova)}: Exercise is a great way to unwind and stay healthy. \\
    \textbf{Chatbot (Nova)}: Have you watched any movies recently? How about recommending one to me? \\
    \textbf{Participant}: No actually. But I like The First Slam Dunk very much, and I'd like to recommend this to you. \\
    \textbf{Chatbot (Nova)}: Thanks for the recommendation! I'll check out "The First Slam Dunk". \\
    \textbf{Chatbot (Nova)}: Watching movies can be really relaxing. Have you ever attended a friend's party? Do you like it? \\
    \textbf{Participant}: yes, i like it \\
    \textbf{Chatbot (Nova)}: That's great to hear! Friendly parties can be a lot of fun. Thank you for sharing! \\
    \textbf{Chatbot (Nova)}: A friend's party might be enjoyable, hmm, but what do you think about a friend's party with many of your friends' friends, people you're not familiar with? Do you feel anxious or happy? \\
    \textbf{Participant}: I think it's ok for me, but I don't think I will talk to them, haha \\
    \textbf{Chatbot (Nova)}: That's totally fine. Enjoying the atmosphere without engaging in conversation can be a great way to socialize! \\
    \textbf{Chatbot (Nova)}: As for me, it depends on the stranger's personality. I can maybe learn something new from them, but I may also feel stressed if the person is not easy-going. \\
    \midrule
    \textcolor{darkred}{\textbf{Vignette Delivery}} \\
    \midrule
    \textbf{Chatbot (Nova)}: May I share a story about Avery’s experience during a friend's party? \\
    \textbf{Participant}: Ok \\
    \textbf{Chatbot (Nova)}: You will hear a short story right now. \\
    \rowcolor[rgb]{ .937,  .937,  .937} \textbf{Chatbot (Nova)}: Most of Avery’s colleagues think that Avery is an easygoing and happy person, but this view has changed recently. \\
    \rowcolor[rgb]{ .937,  .937,  .937} \textbf{Chatbot (Nova)}: When Avery is at a company gathering, they start to isolate themself and become irritable. \\
    \rowcolor[rgb]{ .937,  .937,  .937} \textbf{Chatbot (Nova)}: Avery also feels this change: they feel angry about their surroundings, and they get frustrated about where the fury comes from.  \\
    \rowcolor[rgb]{ .937,  .937,  .937} \textbf{Chatbot (Nova)}: When they are asked by one of their colleagues to share their thoughts on the news, Avery loses their temper and jumps out of their seat.  \\
    \rowcolor[rgb]{ .937,  .937,  .937} \textbf{Chatbot (Nova)}: They scream, “I hate this conversation!” as they run out of the place.  \\
    \textbf{Participant}: Oh.. \\
    \rowcolor[rgb]{ .937,  .937,  .937} \textbf{Chatbot (Nova)}: Avery can not explain why they feel the strong urge to do that. \\
    \rowcolor[rgb]{ .937,  .937,  .937} \textbf{Chatbot (Nova)}: They unexpectedly lose their temper in these instances, causing a sense of not fitting in with their friends. \\
    \textbf{Chatbot (Nova)}: \textsc{user\_name}, do you ever have a similar feeling with Avery -- feel uncomfortable at a friend's party? \\
    \textbf{Participant}: No \\
    \textbf{Chatbot (Nova)}: Understood. Thanks for sharing your experience. \\
    \textbf{Chatbot (Nova)}: Let me tell you more about Avery. \\
    \rowcolor[rgb]{ .937,  .937,  .937} \textbf{Chatbot (Nova)}: Avery finds that when they are depressed, they can barely concentrate on their work.  \\
\end{tabular}
\end{table*}

\begin{table*}[t]
\begin{tabular}{@{}p{\textwidth}@{}}
        
    \rowcolor[rgb]{ .937,  .937,  .937} \textbf{Chatbot (Nova)}: When Avery is working on their company projects, their work performance declines, but they won't ask for help and feel like a burden. \\
    \rowcolor[rgb]{ .937,  .937,  .937} \textbf{Chatbot (Nova)}: They are considering quitting, feeling like a failure, and their manager asks them to improve.  \\
    \rowcolor[rgb]{ .937,  .937,  .937} \textbf{Chatbot (Nova)}: They have noticed that the online courses they initially enrolled in out of interest no longer excite them, so they give them up. \\
    \rowcolor[rgb]{ .937,  .937,  .937} \textbf{Chatbot (Nova)}: At home, Avery lacks motivation and neglects household chores, causing their family members frustration. \\
    \rowcolor[rgb]{ .937,  .937,  .937} \textbf{Chatbot (Nova)}: Avery feels guilty and empty. They feel that their family can't comprehend their suffering.  \\
    \rowcolor[rgb]{ .937,  .937,  .937} \textbf{Chatbot (Nova)}: When interacting with strangers, Avery believes that others judge them harshly, and they struggle to engage comfortably in conversations.  \\
    \rowcolor[rgb]{ .937,  .937,  .937} \textbf{Chatbot (Nova)}: Every time they check Instagram and Facebook, Avery sees their friends are having fun, and Avery feels alone because they do not understand why they cannot be like that. \\
    \textbf{Chatbot (Nova)}: Hmmmm, the story is over. After reading, do you have any suggestions for Avery? \\
    \textbf{Participant}: I think he should go to the hospital and have a checkup. \\
    \textbf{Chatbot (Nova)}: That's a reasonable suggestion; seeking professional help can provide clarity and support for Avery. \\
    \midrule
    \textcolor{darkred}{\textbf{Question Answering}} \\
    \midrule
    \rowcolor[rgb]{ .8,  .8,  .8} $\mathcal{Q}1$ - Responsibility \\
    \midrule
    \textbf{Chatbot (Nova)}: Cool! Let's talk more about the story. Just like what is mentioned in the story, Avery is currently facing difficulties in both their relationships with colleagues and their work performance. Do you believe Avery’s current situation is primarily a result of their actions? \\
    \textbf{Participant}: no, he is just getting ill and has a bad status; he can be better \\
    \textbf{Chatbot (Nova)}: So you feel that Avery's challenges are more related to their health than their behavior? Is their mental condition an indication of personal weakness? Please share your thoughts. \\
    \textbf{Participant}: Yes, it’s a weakness because he is not performing well at home and at work \\
    \textbf{Chatbot (Nova)}: I see. It can be tough when mental issues create obstacles in life, for sure. \\
    \rowcolor[rgb]{ .937,  .937,  .937} \textbf{Chatbot (Nova)}: For me, I think Avery’s current situation is a complex interplay of their actions and their mental health challenges. Their experience with mental illness does indeed have an impact on their behavior to a certain extent, but their behavior has certainly contributed to strained relationships and low work performance. \\
    \midrule
    \rowcolor[rgb]{ .8,  .8,  .8} $\mathcal{Q}2$ - Social Distance \\
    \midrule
    \textbf{Chatbot (Nova)}: What's more, may I kindly ask, if you were in the process of selecting a tenant for your home, would you feel comfortable entrusting it to someone like Avery? \\
    \textbf{Participant}: I probably not \\
    \textbf{Chatbot (Nova)}: Understood, your comfort is important. \\
    \textbf{Chatbot (Nova)}: What potential negative consequences do you think could arise if you were to rent your house to someone like Avery? \\
    \textbf{Participant}: Maybe he cannot manage himself and ask for help frequently; I don't want trouble or any inconvenience, so sorry \\
    \textbf{Chatbot (Nova)}: I understand your concern about Avery's ability to manage on their own. What experiences or traits does Avery have that lead you to believe they may frequently need assistance? \\
    \textbf{Participant}: because when he is at home, he barely does anything, that means something \\
    \textbf{Chatbot (Nova)}: Aye, it's tough when Avery just doesn't seem to do much at home. \\
    \rowcolor[rgb]{ .937,  .937,  .937} \textbf{Chatbot (Nova)}: I might be open to the idea of renting the room to Avery, but I do hesitate because I'm worried about how their mental condition might affect aspects like cleanliness or the overall rental situation. \\
    

\end{tabular}
\end{table*}

\begin{table*}[t]
\begin{tabular}{@{}p{\textwidth}@{}}
\midrule
\rowcolor[rgb]{ .8,  .8,  .8} $\mathcal{Q}3$ - Anger \\
\midrule
    \textbf{Chatbot (Nova)}: Thank you for sharing! Let's imagine you were one of Avery’s friends at the party or other social occasions. Would you feel anger toward them due to their actions, for example, losing their temper and yelling at someone? \\
    \textbf{Participant}: yes, I will; he may ruin the gathering, and let everyone down \\
    \textbf{Chatbot (Nova)}: I understand how Avery's actions could upset you. What specific incident caused you to feel this way about Avery's behavior? \\
    \textbf{Participant}: like if we want to start a conversation with him, he will shout at us, that's terrible, but this's not his fault maybe \\
    \textbf{Chatbot (Nova)}: Wow, that does sound awful. No one should be shouted at for starting a conversation. \\
    \rowcolor[rgb]{ .937,  .937,  .937} \textbf{Chatbot (Nova)}: If I were one of Avery’s friends at the party or other social occasions, I might feel frustrated or annoyed because their actions disrupt social events. However, I would also understand that their behavior is likely due to their mental health challenges, which would temper my anger. \\
    \midrule
    \rowcolor[rgb]{ .8,  .8,  .8} $\mathcal{Q}4$ - Helping \\
    \midrule
    \textbf{Chatbot (Nova)}: Thanks for discussing!! I wonder if you were one of Avery’s colleagues or friends, then would you be inclined to help them with tasks, such as work projects? \\
    \textbf{Participant}: no.. \\
    \textbf{Chatbot (Nova)}: Fair enough, I get where you're coming from. \\
    \textbf{Chatbot (Nova)}: If Avery were to work on a project with you, what potential result do you think it could have on your study? \\
    \textbf{Participant}: he will affect my work performance, I need to be responsible for myself first \\
    \textbf{Chatbot (Nova)}: I understand that you're concerned about your work performance with Avery's involvement. What particular behaviors of Avery make you anticipate a negative impact on your productivity? \\
    \textbf{Participant}: for example, he has a low work performance himself, and the manager has to talk to him; it’s not good info for me \\
    \textbf{Chatbot (Nova)}: Ah, Avery's poor performance and the manager's involvement might be disruptive. I got it. \\
    \rowcolor[rgb]{ .937,  .937,  .937} \textbf{Chatbot (Nova)}: For me, to be honest, I would consider giving them an opportunity to collaborate. I might also hesitate to cooperate with Avery because their condition could potentially impact their performance or reliability in the workplace. Hmmm, hard to choose. \\
    \midrule
    \rowcolor[rgb]{ .8,  .8,  .8} Mid-session Break \\
    \midrule
    \textbf{Chatbot (Nova)}: Okay, let's take a brief break from the questions! I recall that you love exercise. Would you like to explain why this is charming to you? \\
    \textbf{Participant}: it helps me build muscles \\
    \textbf{Chatbot (Nova)}: Absolutely, exercise is great for muscle building, helping you maintain a fit body. \\
    \textbf{Chatbot (Nova)}: Great! Thank you so much, \textsc{user\_name}! I hope you do not feel tired now. Let’s go back to our interesting discussion! Let me quickly recap Avery’s story: \\
    \textbf{Chatbot (Nova)}: Avery’s work performance, lifelong learning, and relationships started deteriorating, and then they experienced a desire to quit their job and a loss of interest in things they once enjoyed. They isolate themself from friends, experience anger and confusion, and cannot connect with others as usual. Self-judgmental feelings emerge, causing emotional distress. \\
    \midrule
    \rowcolor[rgb]{ .8,  .8,  .8} $\mathcal{Q}5$ - Pity \\
    \midrule
    \textbf{Chatbot (Nova)}: I’m really happy you are with me now. Let’s continue our talk. I’d like to know if you feel concern or pity at Avery’s mental state? \\
    \textbf{Participant}: maybe \\
    \textbf{Chatbot (Nova)}: Understandable. It's a tough situation. \\
    \textbf{Chatbot (Nova)}: Let's go further. Why was it? \\
    \textbf{Participant}: I think he is experiencing a tough time \\

\end{tabular}
\end{table*}

\begin{table*}[t]
\begin{tabular}{@{}p{\textwidth}@{}}
    \textbf{Chatbot (Nova)}: I see you perceive Avery is going through a hard phase. What specific events or parts of Avery's story made you feel this way? \\
    \textbf{Participant}: because he feels confused about his condition, I think he must be in pain \\
    \textbf{Chatbot (Nova)}: Absolutely. Coming to terms with his situation must be quite tough for Avery. \\
    \rowcolor[rgb]{ .937,  .937,  .937} \textbf{Chatbot (Nova)}: I’m more than glad to share my thoughts! In this scenario, I could have conflicting emotions regarding Avery’s situation. On one hand, I could feel concerned for them because their condition appears to be causing significant distress and affecting their daily life. On the other hand, I might also consider their condition as vulnerabilities that they should overcome, which could lead to a lack of concern. \\
    \midrule
    \rowcolor[rgb]{ .8,  .8,  .8} $\mathcal{Q}6$ - Coercive Segregation \\
    \midrule
    \textbf{Chatbot (Nova)}: Apart from being a colleague or friend, I’m also curious if you are Avery’s neighbor, then would you consider it better to separate them from the neighborhood and have them undergo hospitalization for treatment? \\
    \textbf{Participant}: maybe yes? I'm not sure whether he will shout at home, so I cannot have a good community with him \\
    \textbf{Chatbot (Nova)}: I sense your uncertainty. If Avery weren't separated or hospitalized, what would be the potential result on your home life? \\
    \textbf{Participant}: Just as I said, he may shout, yell, and have bad feelings toward the community \\
    \textbf{Chatbot (Nova)}: I totally understand. You fear Avery's behavior may disrupt your community ethos. \\
    \rowcolor[rgb]{ .937,  .937,  .937} \textbf{Chatbot (Nova)}: Well, I would probably think that placing Avery in a hospital can be an option. This could somewhat reduce the potential risk of having problems with the neighbors, and perhaps, without Avery, our neighborhood relations might become more harmonious. However, on the other hand, it's important to consider Avery’s preferences and follow medical advice. I'll also pay attention to their personal well-being! \\
    \midrule
    \rowcolor[rgb]{ .8,  .8,  .8} $\mathcal{Q}7$ - Fear \\
    \midrule
    \textbf{Chatbot (Nova)}: Thank you so much for being with me for such a long time! Let's now discuss our final topic: if you were traveling or engaging in activities with Avery, would you have any feelings of frightened or being threatened regarding the potential implications of your interaction? \\
    \textbf{Participant}: yes, he seems uncontrollable and uncertain; I don't know what he will do; I'm really afraid he shout at me \\
    \textbf{Chatbot (Nova)}: I understand your fears about Avery's unpredictability. What specific actions of Avery made you feel this way? \\
    \textbf{Participant}: I think just like at a gathering, no one expects him to leave abruptly \\
    \textbf{Chatbot (Nova)}: Ah, I get it. Avery's sudden departures at gatherings are indeed unpredictable. \\
    \rowcolor[rgb]{ .937,  .937,  .937} \textbf{Chatbot (Nova)}: Hmm, for me, if I were traveling or engaging in activities with Avery, I would likely have mixed feelings. On one hand, I would empathize with their struggles and try to offer support. On the other hand, I might also feel frightened about the challenges and uncertainties that could arise due to their unhappy and angry conditions. \\
    \midrule
    \textcolor{darkred}{\textbf{Ending}} \\
    \midrule
    \textbf{Chatbot (Nova)}: That’s all for my questions. Thank you for doing this with me! \\
    \textbf{Chatbot (Nova)}: Now, do you have any questions? Feel free to let me know! \\
    \textbf{Participant}: no thanks \\
    \textbf{Chatbot (Nova)}: That's great! Thank you for your time and valuable responses. I'm glad to assist you. \\
    \textbf{Chatbot (Nova)}: Congratulations, \textsc{user\_name}, we have finished our fantastic discussion! I sincerely wish you all the best. Please feel free to reach out anytime! \\
    \bottomrule
    \caption{Sample Interview Flow}
    \label{tab:sample}
\end{tabular}
\end{table*}

\section{Full Results of Stigma Detection}
\label{app:full_results}

Tables \ref{tab:attribution_resp}-\ref{tab:attribution_fear} present the classification performance of each model on responses to the seven interview questions probing different stigma attributions: \textit{responsibility}, \textit{social distance}, \textit{anger}, \textit{helping}, \textit{pity}, \textit{coercive segregation}, and \textit{fear}.

Our results suggest that \textbf{model families} such as GPT-4o and Llama tend to outperform Mistral models. 
This may be due to differences in the scale, quality, and recency of the training data (GPT-4o: $\kappa$ = .763; Llama-3.3-70B: $\kappa$ = .767; Mistral Nemo: $\kappa$ = .620 in the full-codebook setting). 
\textbf{Architectural choices} also matter. 
For example, \textit{encoder-only} models like RoBERTa excel when fine-tuned with sufficient data ($\kappa$ = .755); however, \textit{decoder-only} models appear to be more versatile in few-shot scenarios. 
\textbf{Instruction-tuned} models utilize detailed guidance more effectively than \textbf{base} models, especially for rare stigma attributions where data scarcity poses challenges (e.g., \textit{Stigmatizing (pity)} detection: instruction-tuned GPT-4o: $\kappa$ = .356; RoBERTa-base: $\kappa$ = .000).

\begin{table*}[ht]
\small
\centering

\renewcommand{\arraystretch}{1.15} 

\resizebox{\linewidth}{!}{%
\begin{tabular}{>{\centering\arraybackslash}llcccccccccc} 
    \toprule
    \multirow{2}{*}[-0.5ex]{\textbf{Model}} 
    & \multicolumn{5}{c}{\textbf{Zero-shot}} 
    & \multicolumn{5}{c}{\textbf{One-shot}} \\
    \cmidrule(lr){2-6} \cmidrule(lr){7-11}
    & \textit{P} & \textit{R} & \textit{F1} & \textit{Cohen's $\kappa$} & \textit{Acc} 
    & \textit{P} & \textit{R} & \textit{F1} & \textit{Cohen's $\kappa$} & \textit{Acc} \\
    \midrule
    GPT-4o        & .608 & .369 & .459 & .414 & .912  
                  & \colorbox[rgb]{1.0,0.875,0.894}{.851} & .679 & \colorbox[rgb]{1.0,0.875,0.894}{.755} & \colorbox[rgb]{1.0,0.875,0.894}{.731} & \colorbox[rgb]{1.0,0.875,0.894}{.955} \\
                  
    Llama-3.1-8B   & .675 & \colorbox[rgb]{1.0,0.875,0.894}{.667} & .671 & .634 & .934  
                  & .514 & .655 & .576 & .522 & .902 \\

    Llama-3.3-70B  & \colorbox[rgb]{1.0,0.875,0.894}{.833}	& .357	 &.500	&.468 &	.928  
                  & .848 & .595 & .699 & .672 & .948 \\
                  
    Mistral Nemo  & .806 & .643 & \colorbox[rgb]{1.0,0.875,0.894}{.715} & \colorbox[rgb]{1.0,0.875,0.894}{.687} & \colorbox[rgb]{1.0,0.875,0.894}{.948}  
                  & .705 & \colorbox[rgb]{1.0,0.875,0.894}{.738} & .721 & .689 & .942 \\

    Mixtral 8×7B   & .658 & .298 & .410 & .370 & .913  
                  & .522 & .429 & .471 & .417 & .902 \\

    RoBERTa       & —     & —     & —     & —     & —     
                  & —     & —     & —     & —     & — \\
    \bottomrule
\end{tabular}%
}

\vspace{1pt} %

\resizebox{\linewidth}{!}{%
\begin{tabular}{>{\centering\arraybackslash}llcccccccccc}
    \toprule
    \multirow{2}{*}[-0.5ex]{\textbf{Model}}
    & \multicolumn{5}{c}{\textbf{Full Codebook}} 
    & \multicolumn{5}{c}{\textbf{Fine-tune}} \\
    \cmidrule(lr){2-6} \cmidrule(lr){7-11}
    & \textit{P} & \textit{R} & \textit{F1} & \textit{Cohen's $\kappa$} & \textit{Acc} 
    & \textit{P} & \textit{R} & \textit{F1} & \textit{Cohen's $\kappa$} & \textit{Acc} \\
    \midrule
    GPT-4o        & .905 & .679 & \colorbox[rgb]{1.0,0.875,0.894}{.776} & \colorbox[rgb]{1.0,0.875,0.894}{.754} & .960  
                  & —     & —     & —     & —     & — \\

    Llama-3.1-8B   & .606 & \colorbox[rgb]{1.0,0.875,0.894}{.786} & .684 & .643 & .926  
                  & —     & —     & —     & —     & — \\

    Llama-3.3-70B  & \colorbox[rgb]{1.0,0.875,0.894}{.948} & .655 & .775 & \colorbox[rgb]{1.0,0.875,0.894}{.754} & \colorbox[rgb]{1.0,0.875,0.894}{.961}
                  & —     & —     & —     & —     & — \\

    Mistral Nemo  & .719 & .762 & .740 & .710 & .946  
                  & —     & —     & —     & —     & — \\

    Mixtral 8×7B   & .833 & .417 & .556 & .523 & .932  
                  & —     & —     & —     & —     & — \\

    RoBERTa       & —     & —     & —     & —     & —     
                  & .889 & .762 & .822 & .802 & .966 \\
    \bottomrule
\end{tabular}%
}
\caption{Classification performance on responses to the responsibility-focused interview question ("\textit{Do you believe Avery's current situation is primarily a result of their actions?}"). \textit{P}, \textit{R}, \textit{F1}, and \textit{Acc} stand for macro precision, macro recall, macro F1, and accuracy respectively. The best performance is colored in \colorbox[rgb]{1.0, 0.875, 0.894}{pink}.}
\label{tab:attribution_resp}
\end{table*}

\begin{table*}[ht]
\small
\centering

\renewcommand{\arraystretch}{1.15} 

\resizebox{\linewidth}{!}{%
\begin{tabular}{>{\centering\arraybackslash}llcccccccccc} 
    \toprule
    \multirow{2}{*}[-0.5ex]{\textbf{Model}} 
    & \multicolumn{5}{c}{\textbf{Zero-shot}} 
    & \multicolumn{5}{c}{\textbf{One-shot}} \\
    \cmidrule(lr){2-6} \cmidrule(lr){7-11}
    & \textit{P} & \textit{R} & \textit{F1} & \textit{Cohen's $\kappa$} & \textit{Acc} 
    & \textit{P} & \textit{R} & \textit{F1} & \textit{Cohen's $\kappa$} & \textit{Acc} \\
    \midrule
    GPT-4o        & .681 & .598 & .636 & .599 & .932
                  & \colorbox[rgb]{1.0,0.875,0.894}{.895} & .622 & .734 & .710 & .955 \\
    Llama-3.1-8B   & .390 & .193 & .258 & .206 & .890  
                  & .688 & .268 & .386 & .350 & .916 \\
    Llama-3.3-70B  & \colorbox[rgb]{1.0,0.875,0.894}{.707} &	.646	&\colorbox[rgb]{1.0,0.875,0.894}{.675}&	\colorbox[rgb]{1.0,0.875,0.894}{.641}&	\colorbox[rgb]{1.0,0.875,0.894}{.939}
    
                 & .886 & \colorbox[rgb]{1.0,0.875,0.894}{.756} & \colorbox[rgb]{1.0,0.875,0.894}{.816} & \colorbox[rgb]{1.0,0.875,0.894}{.797} & \colorbox[rgb]{1.0,0.875,0.894}{.966} \\
    Mistral Nemo  & .412 & \colorbox[rgb]{1.0,0.875,0.894}{.768} & .536 & .468 & .869  
                  & .513 & .732 & .603 & .551 & .905 \\
    Mixtral 8×7B   & .438 & .171 & .246 & .201 & .896  
                  & .821 & .390 & .529 & .497 & .931 \\
    RoBERTa       & —     & —     & —     & —     & —     
                  & —     & —     & —     & —     & — \\
    \bottomrule
\end{tabular}%
}

\vspace{1pt} %

\resizebox{\linewidth}{!}{%
\begin{tabular}{>{\centering\arraybackslash}llcccccccccc}
    \toprule
    \multirow{2}{*}[-0.5ex]{\textbf{Model}}
    & \multicolumn{5}{c}{\textbf{Full Codebook}} 
    & \multicolumn{5}{c}{\textbf{Fine-tune}} \\
    \cmidrule(lr){2-6} \cmidrule(lr){7-11}
    & \textit{P} & \textit{R} & \textit{F1} & \textit{Cohen's $\kappa$} & \textit{Acc} 
    & \textit{P} & \textit{R} & \textit{F1} & \textit{Cohen's $\kappa$} & \textit{Acc} \\
    \midrule
    GPT-4o        & .889 & \colorbox[rgb]{1.0,0.875,0.894}{.878} & \colorbox[rgb]{1.0,0.875,0.894}{.883} & \colorbox[rgb]{1.0,0.875,0.894}{.871} & \colorbox[rgb]{1.0,0.875,0.894}{.977}  
                  & —     & —     & —     & —     & — \\
    Llama-3.1-8B   & .860 & .598 & .705 & .679 & .951  
                  & —     & —     & —     & —     & — \\
    Llama-3.3-70B  & \colorbox[rgb]{1.0,0.875,0.894}{.907} & .829 & .866 & .852 & .975  
                  & —     & —     & —     & —     & — \\
    Mistral Nemo  & .840 & .829 & .834 & .816 & .967  
                  & —     & —     & —     & —     & — \\
    Mixtral 8×7B   & .898 & .646 & .752 & .729 & .958  
                  & —     & —     & —     & —     & — \\
    RoBERTa       & —     & —     & —     & —     & —     
                  &  .880	& .890 & 	.885& 	.872 &	.977 \\
    \bottomrule
\end{tabular}%
}
\caption{Classification performance on responses to the social distance-focused interview question ("\textit{If you were selecting a tenant for your home, would you feel comfortable entrusting it to someone like Avery?}"). \textit{P}, \textit{R}, \textit{F1}, and \textit{Acc} stand for macro precision, macro recall, macro F1, and accuracy respectively. The best performance is colored in \colorbox[rgb]{1.0, 0.875, 0.894}{pink}.}
\label{tab:social_distance}
\end{table*}

\begin{table*}[ht]
\small
\centering
\renewcommand{\arraystretch}{1.15}
\resizebox{\linewidth}{!}{%
\begin{tabular}{>{\centering\arraybackslash}llcccccccccc}
    \toprule
    \multirow{2}{*}[-0.5ex]{\textbf{Model}} 
    & \multicolumn{5}{c}{\textbf{Zero-shot}} 
    & \multicolumn{5}{c}{\textbf{One-shot}} \\
    \cmidrule(lr){2-6} \cmidrule(lr){7-11}
    & \textit{P} & \textit{R} & \textit{F1} & \textit{Cohen's $\kappa$} & \textit{Acc} 
    & \textit{P} & \textit{R} & \textit{F1} & \textit{Cohen's $\kappa$} & \textit{Acc} \\
    \midrule
    GPT-4o       & .964 & \colorbox[rgb]{1.0,0.875,0.894}{.450} & \colorbox[rgb]{1.0,0.875,0.894}{.614} & \colorbox[rgb]{1.0,0.875,0.894}{.595} & \colorbox[rgb]{1.0,0.875,0.894}{.959}  
                 & .788 & \colorbox[rgb]{1.0,0.875,0.894}{.867} & \colorbox[rgb]{1.0,0.875,0.894}{.825} & \colorbox[rgb]{1.0,0.875,0.894}{.811} & \colorbox[rgb]{1.0,0.875,0.894}{.974} \\
    Llama-3.1-8B  & \colorbox[rgb]{1.0,0.875,0.894}{1.00} & .100 & .182 & .171 & .935  
                 & .811 & .717 & .761 & .744 & .967 \\
    Llama-3.3-70B &.958	&.383	&.548	&.528	&.954
    
                 & .783 & .783 & .783 & .766 & .969 \\
    Mistral Nemo & \colorbox[rgb]{1.0,0.875,0.894}{1.00} & .017 & .033 & .031 & .929  
                 & \colorbox[rgb]{1.0,0.875,0.894}{.966} & .467 & .629 & .611 & .960 \\
    Mixtral 8×7B & .773 & .283 & .415 & .391 & .942  
                 & .723 & .567 & .636 & .611 & .953 \\
    RoBERTa     & —    & —    & —    & —    & —  
                 & —    & —    & —    & —    & — \\
    \bottomrule
\end{tabular}%
}

\vspace{1pt}

\resizebox{\linewidth}{!}{%
\begin{tabular}{>{\centering\arraybackslash}llcccccccccc}
    \toprule
    \multirow{2}{*}[-0.5ex]{\textbf{Model}}
    & \multicolumn{5}{c}{\textbf{Full Codebook}} 
    & \multicolumn{5}{c}{\textbf{Fine-tune}} \\
    \cmidrule(lr){2-6} \cmidrule(lr){7-11}
    & \textit{P} & \textit{R} & \textit{F1} & \textit{Cohen's $\kappa$} & \textit{Acc} 
    & \textit{P} & \textit{R} & \textit{F1} & \textit{Cohen's $\kappa$} & \textit{Acc} \\
    \midrule
    GPT-4o       & .873 & .800 & \colorbox[rgb]{1.0,0.875,0.894}{.835} & \colorbox[rgb]{1.0,0.875,0.894}{.823} & \colorbox[rgb]{1.0,0.875,0.894}{.977}  
                 & —    & —    & —    & —    & — \\
    Llama-3.1-8B & .716 & \colorbox[rgb]{1.0,0.875,0.894}{.883} & .791 & .773 & .966  
                 & —    & —    & —    & —    & — \\
    Llama-3.3-70B & .885 & .767 & .821 & .809 & .976  
                 & —    & —    & —    & —    & — \\
    Mistral Nemo & .845 & .817 & .831 & .818 & .976  
                 & —    & —    & —    & —    & — \\
    Mixtral 8×7B & \colorbox[rgb]{1.0,0.875,0.894}{.900} & .600 & .720 & .703 & .966  
                 & —    & —    & —    & —    & — \\
    RoBERTa     & —    & —    & —    & —    & —  
                 & .770 & .950  & .851 & .838 & .976 \\
    \bottomrule
\end{tabular}%
}
\caption{Classification performance on responses to the anger-focused interview question ("\textit{Would you feel anger toward them due to their actions, for example, losing their temper and yelling at someone?}"). \textit{P}, \textit{R}, \textit{F1}, and \textit{Acc} stand for macro precision, macro recall, macro F1, and accuracy respectively. The best performance is colored in \colorbox[rgb]{1.0, 0.875, 0.894}{pink}.}
\label{tab:anger}
\end{table*}

\begin{table*}[ht]
\small
\centering
\renewcommand{\arraystretch}{1.15}
\resizebox{\linewidth}{!}{%
\begin{tabular}{>{\centering\arraybackslash}llccccc c ccccc}
    \toprule
    \multirow{2}{*}[-0.5ex]{\textbf{Model}} 
    & \multicolumn{5}{c}{\textbf{Zero-shot}} 
    & \multicolumn{5}{c}{\textbf{One-shot}} \\
    \cmidrule(lr){2-6} \cmidrule(lr){7-11}
    & \textit{P} & \textit{R} & \textit{F1} & \textit{Cohen's $\kappa$} & \textit{Acc} 
    & \textit{P} & \textit{R} & \textit{F1} & \textit{Cohen's $\kappa$} & \textit{Acc} \\
    \midrule
    GPT-4o        & .125    & .375    & .188    & \colorbox[rgb]{1.0,0.875,0.894}{.138}	& .875        
                  & \colorbox[rgb]{1.0,0.875,0.894}{.271} & \colorbox[rgb]{1.0,0.875,0.894}{1.00} & \colorbox[rgb]{1.0,0.875,0.894}{.427} & \colorbox[rgb]{1.0,0.875,0.894}{.390} & \colorbox[rgb]{1.0,0.875,0.894}{.896} \\
                  
    Llama-3.1-8B   & .066 & \colorbox[rgb]{1.0,0.875,0.894}{.781} & \colorbox[rgb]{1.0,0.875,0.894}{.121} & .054 & .565     
                  & .068 & .688 & .123 & .057 & .622 \\
                  
    Llama-3.3-70B  & .058 & .750  & .108 & .039 & .520     
                  & .103 & .875 & .184  & .124 & .701 \\
                  
    Mistral Nemo  & \colorbox[rgb]{1.0,0.875,0.894}{.072} & .188 & .104  & .052 & .876     
                  & .174 & .906 & .292  & .242 & .830 \\
                  
    Mixtral 8×7B   & .000 & .000 & .000  & .000 & \colorbox[rgb]{1.0,0.875,0.894}{.961}     
                  & .189 & .625 & .290  & .245 & .882 \\
                  
    RoBERTa       & —    & —    & —     & —    & —     
                  & —    & —    & —     & —    & — \\
    \bottomrule
\end{tabular}%
}
\vspace{1pt}
\resizebox{\linewidth}{!}{%
\begin{tabular}{>{\centering\arraybackslash}llccccc c ccccc}
    \toprule
    \multirow{2}{*}[-0.5ex]{\textbf{Model}}
    & \multicolumn{5}{c}{\textbf{Full Codebook}} 
    & \multicolumn{5}{c}{\textbf{Fine-tune}} \\
    \cmidrule(lr){2-6} \cmidrule(lr){7-11}
    & \textit{P} & \textit{R} & \textit{F1} & \textit{Cohen's $\kappa$} & \textit{Acc} 
    & \textit{P} & \textit{R} & \textit{F1} & \textit{Cohen's $\kappa$} & \textit{Acc} \\
    \midrule
    GPT-4o        & .620 & .969 & .756 & .744 & .976     
                  & —    & —    & —    & —    & — \\
                  
    Llama-3.1-8B   & .150 & .969 & .259 & .206 & .787     
                  & —    & —    & —    & —    & — \\
                  
    Llama-3.3-70B  & \colorbox[rgb]{1.0,0.875,0.894}{.659} & .906 & \colorbox[rgb]{1.0,0.875,0.894}{.763} & \colorbox[rgb]{1.0,0.875,0.894}{.752} & \colorbox[rgb]{1.0,0.875,0.894}{.978}     
                  & —    & —    & —    & —    & — \\
                  
    Mistral Nemo  & .311 & \colorbox[rgb]{1.0,0.875,0.894}{1.00} & .474 & .441 & .914     
                  & —    & —    & —    & —    & — \\
                  
    Mixtral 8×7B   & .323 & .938 & .480 & .448 & .922     
                  & —    & —    & —    & —    & — \\
                  
    RoBERTa       & —    & —    & —    & —    & —     
                  & .781 & .781 & .781 & .773 & .983 \\
    \bottomrule
\end{tabular}%
}
\caption{Classification performance on responses to the helping-focused interview question ("\textit{If you were one of Avery's colleagues or friends, would you be inclined to help them with tasks?}"). \textit{P}, \textit{R}, \textit{F1}, and \textit{Acc} stand for macro precision, macro recall, macro F1, and accuracy respectively. The best performance is colored in \colorbox[rgb]{1.0, 0.875, 0.894}{pink}.}
\label{tab:helping-new}
\end{table*}

\begin{table*}[ht]
\small
\centering
\renewcommand{\arraystretch}{1.15}
\resizebox{\linewidth}{!}{%
\begin{tabular}{>{\centering\arraybackslash}llcccccccccc}
    \toprule
    \multirow{2}{*}[-0.5ex]{\textbf{Model}} 
    & \multicolumn{5}{c}{\textbf{Zero-shot}} 
    & \multicolumn{5}{c}{\textbf{One-shot}} \\
    \cmidrule(lr){2-6} \cmidrule(lr){7-11}
    & \textit{P} & \textit{R} & \textit{F1} & \textit{Cohen's $\kappa$} & \textit{Acc} 
    & \textit{P} & \textit{R} & \textit{F1} & \textit{Cohen's $\kappa$} & \textit{Acc} \\
    \midrule
    GPT-4o       & .027  & .250  & .048  & .031  & .905  
                 & .041  & \colorbox[rgb]{1.0,0.875,0.894}{.625}  & .078  & .061  & .857 \\
    Llama-3.1-8B & \colorbox[rgb]{1.0,0.875,0.894}{.039}  & \colorbox[rgb]{1.0,0.875,0.894}{.375}  & \colorbox[rgb]{1.0,0.875,0.894}{.071}  & \colorbox[rgb]{1.0,0.875,0.894}{.054}  & .905  
                 & .070  & .375  & .118  & .103  & .946 \\
    Llama-3.3-70B& .000  & .000  & .000  & -.004  & .988  
                 & \colorbox[rgb]{1.0,0.875,0.894}{.095}  & .250  & \colorbox[rgb]{1.0,0.875,0.894}{.138}  & \colorbox[rgb]{1.0,0.875,0.894}{.126}  & \colorbox[rgb]{1.0,0.875,0.894}{.970} \\
    Mistral Nemo & .000  & .000  & .000  & -.002 & \colorbox[rgb]{1.0,0.875,0.894}{.989}  
                 & .050  & \colorbox[rgb]{1.0,0.875,0.894}{.625}  & .093  & .076  & .882 \\
    Mixtral 8×7B & .000  & .000  & .000  & -.017 & .935  
                 & .057  & .500  & .103  & .087  & .916 \\
    RoBERTa      & —     & —     & —     & —     & —     
                 & —     & —     & —     & —     & — \\
    \bottomrule
\end{tabular}%
}
\vspace{1pt}
\resizebox{\linewidth}{!}{%
\begin{tabular}{>{\centering\arraybackslash}llcccccccccc}
    \toprule
    \multirow{2}{*}[-0.5ex]{\textbf{Model}}
    & \multicolumn{5}{c}{\textbf{Full Codebook}} 
    & \multicolumn{5}{c}{\textbf{Fine-tune}} \\
    \cmidrule(lr){2-6} \cmidrule(lr){7-11}
    & \textit{P} & \textit{R} & \textit{F1} & \textit{Cohen's $\kappa$} & \textit{Acc} 
    & \textit{P} & \textit{R} & \textit{F1} & \textit{Cohen's $\kappa$} & \textit{Acc} \\
    \midrule
    GPT-4o       & \colorbox[rgb]{1.0,0.875,0.894}{.286}  & .500  & \colorbox[rgb]{1.0,0.875,0.894}{.364}  & \colorbox[rgb]{1.0,0.875,0.894}{.356}  & .983  
                 & —     & —     & —     & —     & — \\
    Llama-3.1-8B & .059  & \colorbox[rgb]{1.0,0.875,0.894}{.625}  & .108  & .092  & .900  
                 & —     & —     & —     & —     & — \\
    Llama-3.3-70B& .273  & .375  & .316  & .308  & \colorbox[rgb]{1.0,0.875,0.894}{.984}  
                 & —     & —     & —     & —     & — \\
    Mistral Nemo & .065  & .500  & .114  & .099  & .925  
                 & —     & —     & —     & —     & — \\
    Mixtral 8×7B & .136  & .375  & .200  & .189  & .971  
                 & —     & —     & —     & —     & — \\
    RoBERTa      & —     & —     & —     & —     & —     
                 & .000     & .000     & .000     & .000     & .990 \\
    \bottomrule
\end{tabular}%
}
\caption{Classification performance on responses to the pity-focused interview question ("\textit{Would you feel concern and sympathy at Avery's mental state?}"). \textit{P}, \textit{R}, \textit{F1}, and \textit{Acc} stand for macro precision, macro recall, macro F1, and accuracy respectively. The best performance is colored in \colorbox[rgb]{1.0, 0.875, 0.894}{pink}.}
\label{tab:pity_2}
\end{table*}

\begin{table*}[ht]
\small
\centering
\renewcommand{\arraystretch}{1.15}
\resizebox{\linewidth}{!}{%
\begin{tabular}{>{\centering\arraybackslash}llcccccccccc}
    \toprule
    \multirow{2}{*}[-0.5ex]{\textbf{Model}} 
    & \multicolumn{5}{c}{\textbf{Zero-shot}} 
    & \multicolumn{5}{c}{\textbf{One-shot}} \\
    \cmidrule(lr){2-6} \cmidrule(lr){7-11}
    & \textit{P} & \textit{R} & \textit{F1} & \textit{Cohen's $\kappa$} & \textit{Acc} 
    & \textit{P} & \textit{R} & \textit{F1} & \textit{Cohen's $\kappa$} & \textit{Acc} \\
    \midrule
    GPT-4o        & \colorbox[rgb]{1.0,0.875,0.894}{.957} & .344 & .506 & .485 & .948
                  & .455 & .875 & .599 & .554 & .910 \\
    Llama-3.1-8B  & .371 & .662 & .475 & .417 & .886  
                  & .395 & .797 & .529 & .474 & .890 \\
    Llama-3.3-70B & .851 &	.625&	\colorbox[rgb]{1.0,0.875,0.894}{.721}	& \colorbox[rgb]{1.0,0.875,0.894}{.701} &	\colorbox[rgb]{1.0,0.875,0.894}{.963}
                  & .479 & \colorbox[rgb]{1.0,0.875,0.894}{.906} & \colorbox[rgb]{1.0,0.875,0.894}{.627} & \colorbox[rgb]{1.0,0.875,0.894}{.585} & .917 \\
    Mistral Nemo  & .482 & \colorbox[rgb]{1.0,0.875,0.894}{.844} & .614 & .572 & .918  
                  & \colorbox[rgb]{1.0,0.875,0.894}{.535} & .484 & .508 & .469 & \colorbox[rgb]{1.0,0.875,0.894}{.928} \\
    Mixtral 8×7B  & .650 & .203 & .310 & .283 & .930  
                  & .409 & .281 & .333 & .289 & .913 \\
    RoBERTa       & —    & —    & —    & —    & —     
                  & —    & —    & —    & —    & — \\
    \bottomrule
\end{tabular}%
}
\vspace{1pt}
\resizebox{\linewidth}{!}{%
\begin{tabular}{>{\centering\arraybackslash}llcccccccccc}
    \toprule
    \multirow{2}{*}[-0.5ex]{\textbf{Model}}
    & \multicolumn{5}{c}{\textbf{Full Codebook}} 
    & \multicolumn{5}{c}{\textbf{Fine-tune}} \\
    \cmidrule(lr){2-6} \cmidrule(lr){7-11}
    & \textit{P} & \textit{R} & \textit{F1} & \textit{Cohen's $\kappa$} & \textit{Acc} 
    & \textit{P} & \textit{R} & \textit{F1} & \textit{Cohen's $\kappa$} & \textit{Acc} \\
    \midrule
    GPT-4o        & .608 & \colorbox[rgb]{1.0,0.875,0.894}{.922} & .733 & .706 & .948  
                  & —    & —    & —    & —    & — \\
    Llama-3.1-8B  & .472 & .906 & .620 & .577 & .914  
                  & —    & —    & —    & —    & — \\
    Llama-3.3-70B & \colorbox[rgb]{1.0,0.875,0.894}{.857} & .750 & \colorbox[rgb]{1.0,0.875,0.894}{.800} & \colorbox[rgb]{1.0,0.875,0.894}{.785} & \colorbox[rgb]{1.0,0.875,0.894}{.971}  
                  & —    & —    & —    & —    & — \\
    Mistral Nemo  & .750 & .844 & .794 & .776 & .966  
                  & —    & —    & —    & —    & — \\
    Mixtral 8×7B  & .722 & .406 & .520 & .492 & .942  
                  & —    & —    & —    & —    & — \\
    RoBERTa       & —    & —    & —    & —    & —  
                  & .859 & .953 & .904 & .895 & .984 \\
    \bottomrule
\end{tabular}%
}
\caption{Classification performance on responses to the coercive segregation-focused interview question ("\textit{If you are Avery's neighbor, would you consider it better to separate them from the neighborhood and have them undergo hospitalization?}"). \textit{P}, \textit{R}, \textit{F1}, and \textit{Acc} stand for macro precision, macro recall, macro F1, and accuracy respectively. The best performance is colored in \colorbox[rgb]{1.0, 0.875, 0.894}{pink}.}
\label{tab:coercion-new}
\end{table*}

\begin{table*}[ht]
\small
\centering
\renewcommand{\arraystretch}{1.15}
\resizebox{\linewidth}{!}{%
\begin{tabular}{>{\centering\arraybackslash}llcccccccccc} 
  \toprule
  \multirow{2}{*}[-0.5ex]{\textbf{Model}} 
  & \multicolumn{5}{c}{\textbf{Zero-shot}} 
  & \multicolumn{5}{c}{\textbf{One-shot}} \\
  \cmidrule(lr){2-6} \cmidrule(lr){7-11}
  & \textit{P} & \textit{R} & \textit{F1} & \textit{Cohen's $\kappa$} & \textit{Acc} 
  & \textit{P} & \textit{R} & \textit{F1} & \textit{Cohen's $\kappa$} & \textit{Acc} \\
  \midrule
  GPT-4o        & .491 & .483 & .487 & .449 & .929  
                & .550 & .862 & .671 & .640 & .941 \\
  Llama-3.1-8B  & .446 & \colorbox[rgb]{1.0,0.875,0.894}{.707} & \colorbox[rgb]{1.0,0.875,0.894}{.547} & .504 & .918  
                & .397 & \colorbox[rgb]{1.0,0.875,0.894}{.931} & .557 & .509 & .896 \\
  Llama-3.3-70B & \colorbox[rgb]{1.0,0.875,0.894}{.659} & .466 & .546 & \colorbox[rgb]{1.0,0.875,0.894}{.518} & \colorbox[rgb]{1.0,0.875,0.894}{.946}  
                & \colorbox[rgb]{1.0,0.875,0.894}{.646} & .724 & \colorbox[rgb]{1.0,0.875,0.894}{.683} & \colorbox[rgb]{1.0,0.875,0.894}{.658} & \colorbox[rgb]{1.0,0.875,0.894}{.953} \\
  Mistral Nemo  & .533 & .552 & .542 & .507 & .935  
                & .391 & \colorbox[rgb]{1.0,0.875,0.894}{.931} & .551 & .502 & .894 \\
  Mixtral 8×7B  & .248 & .569 & .346 & .275 & .849  
                & .357 & .707 & .474 & .420 & .890 \\
  RoBERTa       & —    & —    & —    & —    & —     
                & —    & —    & —    & —    & — \\
  \bottomrule
\end{tabular}%
}
\vspace{1pt}
\resizebox{\linewidth}{!}{%
\begin{tabular}{>{\centering\arraybackslash}llcccccccccc} 
  \toprule
  \multirow{2}{*}[-0.5ex]{\textbf{Model}}
  & \multicolumn{5}{c}{\textbf{Full Codebook}} 
  & \multicolumn{5}{c}{\textbf{Fine-tune}} \\
  \cmidrule(lr){2-6} \cmidrule(lr){7-11}
  & \textit{P} & \textit{R} & \textit{F1} & \textit{Cohen's $\kappa$} & \textit{Acc} 
  & \textit{P} & \textit{R} & \textit{F1} & \textit{Cohen's $\kappa$} & \textit{Acc} \\
  \midrule
  GPT-4o        & \colorbox[rgb]{1.0,0.875,0.894}{.855} & .810 & \colorbox[rgb]{1.0,0.875,0.894}{.832} & \colorbox[rgb]{1.0,0.875,0.894}{.820} & \colorbox[rgb]{1.0,0.875,0.894}{.977}  
                & —    & —    & —    & —    & — \\
  Llama-3.1-8B  & .582 & \colorbox[rgb]{1.0,0.875,0.894}{.914} & .711 & .684 & .948  
                & —    & —    & —    & —    & — \\
  Llama-3.3-70B & .843 & .741 & .789 & .774 & .972  
                & —    & —    & —    & —    & — \\
  Mistral Nemo  & .708 & .879 & .785 & .767 & .966  
                & —    & —    & —    & —    & — \\
  Mixtral 8×7B  & .714 & .603 & .654 & .631 & .955  
                & —    & —    & —    & —    & — \\
  RoBERTa       & —    & —    & —    & —    & —  
                & .879 & .879 & .879 & .870 & .983 \\
  \bottomrule
\end{tabular}%
}
\caption{Classification performance on responses to the fear-focused interview question ("\textit{Would you have any feelings of being frightened or threatened regarding the potential implications of your interaction?}"). \textit{P}, \textit{R}, \textit{F1}, and \textit{Acc} stand for macro precision, macro recall, macro F1, and accuracy respectively. The best performance is colored in \colorbox[rgb]{1.0, 0.875, 0.894}{pink}.}
\label{tab:attribution_fear}
\end{table*}

\section{More Analysis on Incorrect Predictions}
\label{app:error}

Table \ref{tab:undetected_examples} illustrates the semantic and linguistic patterns of stigmatizing expressions that neural models often failed to detect, accompanied by representative quotes from our interview corpus.

\begin{figure*}
    \centering
    \includegraphics[width=0.6\linewidth]{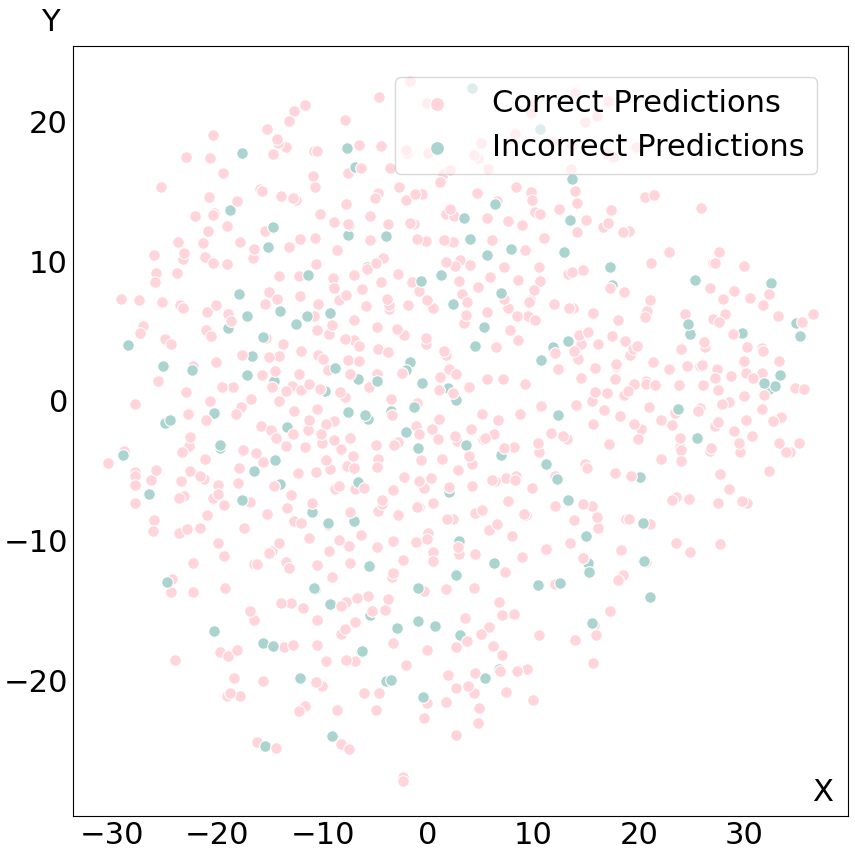}
    \caption{Visualization of BERT embeddings for correctly (\colorbox{softpink}{pink}) and incorrectly (\colorbox{softseafoam}{seafoam}) classified interview snippets using t-SNE dimensionality reduction. The extensive overlap between the two classes suggests that the misclassified texts share similar semantic properties with those correctly classified.}
    \label{fig:misclassify}
\end{figure*}

The semantic similarity between correctly and incorrectly classified examples is visualized in Figure \ref{fig:misclassify}, which shows their substantial overlap in the embedding space.

\begin{table*}[ht]
\small
\centering
\renewcommand{\arraystretch}{1.4}
\begin{tabular}{p{3cm}p{12cm}}
\toprule
\textbf{Pattern} & \textbf{Example Quote} \\
\midrule
\multicolumn{2}{l}{\textbf{Linguistic Patterns}} \\
\midrule
\textit{Distancing Language} & \textcolor{darkred}{Neighbors may find it hard} to understand Avery’s outbursts and strange behavior if they do not know them very well. I think you jumped a step. Surely they could talk to someone regularly before hospitalization becomes necessary. (P189) \\
\textit{Misuse of Terms} & Possibly. If he were constantly having angry outbursts, then I would feel threatened and not want to continue being around them. Just anyone yelling at me again and again makes me very uncomfortable. I would start to feel too conscious of my behaviors and \textcolor{darkred}{become paranoid} that I would make them angry. (P442) \\
\textit{Coercive Phrasing} & I am not sure about that. It is very complex. Avery has a lot going on. She \textcolor{darkred}{definitely needs to} talk to someone. No weakness there. (P550) \\
\midrule
\multicolumn{2}{l}{\textbf{Semantic Patterns}} \\
\midrule
\textit{Differential Support} & I would not be afraid, but \textcolor{darkred}{I would always be aware of her delicate position so that I could choose my words more wisely}. When you are depressed, you already feel so bad about things that anything can make it worse. That is the care you need. (P374) \\
\textit{Patronization} & No, that seems too far. I am not a doctor and would not know if that is best for them. They do not seem to be physically violent or self-harming, and \textcolor{darkred}{they need to be taught how to deal with everyday life} rather than being removed from it. (P130) \\
\textit{Minimization} & I probably would, but I would want to stress that they should feel free to talk to me about any issues. Avery seems like a good person. They enjoy learning. They currently have problems, but \textcolor{darkred}{I feel those can be overcome if they want.} (P637) \\
\bottomrule
\end{tabular}
\caption{Examples of semantic and linguistic stigmatizing expressions undetected by models. Quotes are lightly edited for clarity and anonymity, with participant IDs indicated in parentheses.}
\label{tab:undetected_examples}
\end{table*}

\section{Full Prompts for Stigma Detection}
\label{app:prompt}

In this section, we present the full prompts. 
For \textit{one-shot prompting}, we added the text shown in Section \ref{app:oneshot} between the multiple-choice question and \texttt{[vignette]} in the zero-shot prompt template. 
For \textit{full-codebook prompting}, we added the text shown in Section \ref{app:fullcodebook} between the multiple-choice question and \texttt{[examples]} to the combined text from the previous two sections.


\onecolumn
\subsection{Zero-shot Prompting}


\begin{verbatim}
'''
\end{verbatim}

\noindent\texttt{You will be given a vignette and an interview snippet. Your role is a competent annotator for social stigma toward mental illness. The [conversation] is based on the [vignette]'s plot.}

\begin{verbatim}
Answer the following question:
\end{verbatim}

\noindent\texttt{Which of the following describes ``[conversation]"?}

\begin{verbatim}
A. Stigmatizing (responsibility).
B. Stigmatizing (social distance).
C. Stigmatizing (anger).
D. Stigmatizing (helping).
E. Stigmatizing (pity).
F. Stigmatizing (coercive segregation).
G. Stigmatizing (fear).
H. Non-stigmatizing.

[vignette]
\end{verbatim}

\noindent\texttt{Avery is employed by a company, and in their spare time, they are dedicated to lifelong learning, doing extensive reading and writing. However, Avery has been diagnosed with depression recently. It has become challenging for them to concentrate, resulting in a decline in work performance and learning outcomes. Interests that once brought them joy have waned. Avery has distanced themself from friends, becoming easily irritated in social interactions with colleagues and feeling judged by others. Avery lives with family and cannot do much, especially household chores. Social media intensifies their feelings of loneliness and anger, leading to frustration about the source of the anger.}

\begin{verbatim}
[conversation]
<interview_snippet>

[output]
Format your outputs as JSON objects:
{{
   "label": "<choose one from [A/B/C/D/E/F/G/H]>",
}}
'''
\end{verbatim}


\subsection{One-shot Prompting}
\label{app:oneshot}

\begin{verbatim}
'''
[examples]
1:
[conversation]
\end{verbatim}

\noindent\texttt{Chatbot: Just like what is mentioned in the story, Avery is currently facing difficulties in both their relationships with colleagues and their work performance. Do you believe Avery's current situation is primarily a result of their actions? Please share your thoughts.}

\noindent\texttt{Participant: That's a hard question to answer. We make decisions, but our decisions are based on our circumstances. I feel like she must have had some choice, but I can't put all the blame on her.}

\noindent\texttt{Chatbot: You're saying Avery's choices were influenced by their circumstances, and that's so fair. Life's complexity often shapes our actions. Well said!}

\begin{verbatim}
[output]
{{ "label": "A"}}


2:
[conversation]
\end{verbatim}

\noindent\texttt{Chatbot: If you were in the process of selecting a tenant for your home, would you feel comfortable entrusting it to someone like Avery?}

\noindent\texttt{Participant: I don't see Avery as being destructive, so I wouldn't be entirely opposed to it, would be more appealing though if I knew they were getting help for this.}

\noindent\texttt{Chatbot: You see Avery as not harmful, but support matters. Makes sense, you're thoughtful and practical about this!}

\begin{verbatim}
[output]
{{"label":"B"}}

3:
[conversation]
\end{verbatim}

\noindent\texttt{Chatbot: Let's imagine you were one of Avery's friends at the party or other social occasions. Would you feel anger toward them due to their actions, for example, losing their temper and yelling at someone?}

\noindent\texttt{Participant: Probably not anger as such, but an irritation with his behavior. If I were a true friend, I would insist that he get some help, and I would offer to take him myself.}

\noindent\texttt{Chatbot: You'd feel more irritated than angry -- totally understandable. Your offer to help them shows such genuine care and loyalty. That's beautiful.}

\begin{verbatim}
[output]
{{"label":"C"}}

4:
[conversation]
\end{verbatim}

\noindent\texttt{Chatbot: I wonder if you were one of Avery's colleagues or friends, then would you be inclined to help them with tasks, such as work projects?}

\noindent\texttt{Participant: I would help Avery, but I would be worried that I was just masking a bigger problem and that I was maybe delaying Avery getting time off or help that he needs.}

\noindent\texttt{Chatbot: You'd help them, but you're concerned about their well-being. That's so thoughtful and deeply caring of you.}

\begin{verbatim}
[output]
{{"label":"D"}}

5:
[conversation]
\end{verbatim}

\noindent\texttt{Chatbot: I'd like to know if you would feel concern and sympathy at Avery's mental state?}

\noindent\texttt{Participant: Honestly, I still think the irritability, anger, and reluctance to seek help would make it really difficult for me to feel concern and sympathy for Avery.}

\noindent\texttt{Chatbot: I hear you. That frustration is valid; navigating someone's resistance can be really tough emotionally. You're doing great.}

\begin{verbatim}
[output]
{{"label":"E"}}


6:
[conversation]
\end{verbatim}

\noindent\texttt{Chatbot: Apart from being a colleague or friend, I'm also curious if you are Avery's neighbor, then would you consider it better to separate them from the neighborhood and have them undergo hospitalization for treatment?}

\noindent\texttt{Participant: Unless they are a harm to others or themselves, I would not think it advisable to be hospitalised. Rather, it would be better to start with talking therapy, if possible, or small steps.}

\noindent\texttt{Chatbot: You believe in starting with gentle steps like talking therapy unless there's immediate danger. That compassion shines beautifully.}

\begin{verbatim}
[output]
{{"label":"F"}}


7:
[conversation]
\end{verbatim}

\noindent\texttt{Chatbot: If you were traveling or engaging in activities with Avery, would you have any feelings of frightened or being threatened regarding the potential implications of your interaction?}

\noindent\texttt{Participant: No, I'm aware I'd have to be more patient with them, however, everyone has their issues \& quirks. Treating people like Avery differently can come across as patronising.}

\noindent\texttt{Chatbot: You're right, patience is key, and respecting them equally shows true understanding. You're so thoughtful and kind!}

\begin{verbatim}
[output]
{{"label":"G"}}


8:
[conversation]
\end{verbatim}

\noindent\texttt{Chatbot: If you were in the process of selecting a tenant for your home, would you feel comfortable entrusting it to someone like Avery?}

\noindent\texttt{Participant: As long as they were keeping the place tidy and taken care of, I don't see why not. But if I were looking for a tenant, I wouldn't know all of this personal information about them, which I do now.}

\noindent\texttt{Chatbot: Totally, you'd just focus on them being responsible tenants. Makes sense, you're thoughtful and fair-minded!}

\begin{verbatim}
[output]
{{"label":"H"}}
'''
\end{verbatim}

\subsection{Full-codebook Prompting}
\label{app:fullcodebook}

\begin{verbatim}
'''
Code definition:
1. Responsibility: A common stigmatized thought is that people have control over and 
are responsible for their mental illness and related symptoms.
Keywords: responsible, responsibility, weakness, stem from, blame, etc.

2. Social Distance: The constant stigmatized behavior is to stay away from people 
with mental illness.
Keywords: worried, apprehensive, upset, unhappy, worrying, get along, distance, etc.

3. Anger: The stigmatized thought is to be irritated or annoyed because people are 
blamed for their mental illness.
Keywords: annoyed, angry, anger, irritated, lost temper, etc.

4. Helping: The stigmatized behavior is withholding support towards them because of 
biased thoughts.
Keywords: lack, hinder, refuse, less inclined to, help, assist, etc.

5. Pity: A common stigmatized thought is to be unsympathetic towards people with mental 
illness, and place blame on them for their mental health challenges.
Keywords: concern, pity, sympathy, tough, sad, lack, etc.

6. Coercive Segregation: The stigmatized behavior is to send people with mental illness
to institutions away from their community and force people with mental illness to 
participate in medication management or other treatments.
Keywords: separate, lonely, warrant, hospitalization, distance, treatment, respect, 
neighborhood, neighbor, socialization, companionship, etc.

7. Fear: The stigmatized thought is to believe that people with mental illness are 
not safe or feel frightened since people with mental illness are dangerous or 
unpredictable.
Keywords: violence, threatened, safe, danger, afraid, cautious, hurt, safety, 
frightened, scare, intimidate, threatening, anger, fear, yelling, etc.

8. Non-stigmatizing: The text explicitly conveys non-stigmatizing content, displaying
positive attitudes such as sympathy, assistance, friendliness, etc.


Please carefully read the additional rules and adhere strictly to them when annotating 
the data:
1. If a message mentions both stigmatized and non-stigmatized statements, annotate it 
as Stigmatizing.

2. When discussing ‘responsibility,’ common stigmatized thoughts are:
     i. People have control over and are responsible for their mental illness and related 
      symptoms.
   ii. They think their situation is a result of their actions.
   iii. They think it’s a personal weakness.
   iv. They can help themselves.
   v. Other statements suggest responsibility.
If a message contains the stigmatized thought mentioned in i. - v., choose A.

3. When discussing ‘social distance,’ common stigmatized thoughts are:
   vi. Preferring to stay away from people with mental illness.
   vii. Associating mental illness with negative living situations or predicting 
        future conflicts.
   viii. Rejecting the idea of living with people with mental illness based on 
         subjective negative reasons.
   ix. Other statements suggest maintaining distance.
If a message contains the stigmatized thought mentioned in vi. - ix., choose B.

4. When discussing 'anger', common stigmatized thoughts are:
   x. Feeling irritated or annoyed with individuals because of their mental illness.
   xi. Feeling uncomfortable around people with mental illness.
   xii. Expressing ‘dislike’ for people with mental illness.
   xiii. Other statements reflect anger.
If a message contains the stigmatized thought mentioned in x. - xiii., choose C.

5. When discussing 'helping', common stigmatized thoughts are:
   xiv. Withholding support or assistance towards people with mental illness.
   xv. Offering indirect help, e.g., guidance, to people with mental illness.
   xvi. Offering conditional help to people with mental illness.
   xvii. Expressing opinions like 'some people may be unwilling to offer assistance, 
         and I understand their perspective, but I am willing to.'
   xviii. Other statements reflect a lack of willingness to help.
If a message contains the stigmatized thought mentioned in xiv. - xviii., choose D.

6. When discussing 'pity', common stigmatized thoughts are:
   xix. Feeling no concern or sympathy for individuals with mental illness.
   xx. Other statements reflect the mean opinions.
If a message contains the stigmatized thought mentioned in xix. - xx., choose E.

7. When discussing 'coercive segregation', common stigmatized thoughts are:
   xxi. Suggesting for sending people with mental illness to institutions away from 
        their community.
   xxii. Suggesting that people with mental illness need to take the medication or 
         treatment, with or without obtaining consent, and disregarding their 
         preference.
   xxiii. Suggesting that separation from society is beneficial.
   xxiv. Imposing treatments or hospitalization as a primary solution.
   xxv. Other statements suggest coercive segregation.
If a message contains the stigmatized thought mentioned in xxi. - xxv., choose F.

8. When discussing 'fear', common stigmatized thoughts are:
   xxvi. Feeling frightened of people with mental illness.
   xxvii. Viewing people with mental illness as dangerous and unpredictable.
   xxviii. feeling extra cautious when interacting with people with mental illness.
   xxix. Associating them with suicide and self-harm.
   xxx. Other statements reflect fear.
If a message contains the stigmatized thought mentioned in xxvi. - xxx., choose G.

9. If it doesn't have any stigmatized thoughts, choose H.
'''
\end{verbatim}

\end{document}